%% file: main-en.tex
\documentclass[10pt,a4paper]{article}

\usepackage{caption}
\usepackage{subcaption}
\usepackage{graphicx}
\graphicspath{{figures/}}

\usepackage{amssymb}
\usepackage{amsmath} 

\usepackage{color}
\usepackage{algorithmic}
\usepackage{array}
\usepackage{longtable}
\usepackage{lipsum}
\usepackage{hyperref}
\usepackage{float}
\usepackage{CJKutf8}

\begin{document}

\title{Xuanwu: Evolving General Multimodal Models into an Industrial-Grade Foundation for Content Ecosystems} 
\author{Zhiqian Zhang, Xu Zhao, Xiaoqing Xu, Guangdong Liang,\\Weijia Wang, Xiaolei Lv, Bo Li, Jun Gao\\[0.5em]Computational Intelligence Dept, Hello Group Inc.}
\date{} 
\maketitle 

\begin{abstract}
\input{sections/00-abstract-en}
\end{abstract}

\input{sections/01-introduction-en}
\input{sections/02-architecture-en}
\input{sections/03-training-en}
\input{sections/04-evaluation-en}
\input{sections/05-conclusion-en}

\clearpage

\appendix
\input{sections/07-appendix-en}

\clearpage
\bibliography{bibliography/sample,bibliography/pretrain_appendix}{}
\bibliographystyle{unsrt}

\end{document}

%% file: sections/00-abstract-en.tex
\normalsize
In recent years, multimodal large models have continued to improve on general benchmarks. However, in real-world content moderation and adversarial settings, mainstream models still suffer from degraded generalization and catastrophic forgetting because of limited fine-grained visual perception and insufficient modeling of long-tail noise. In this paper, we present Xuanwu VL-2B as a case study of how general multimodal models can be developed into an industrial-grade foundation model for content ecosystems. The model adopts a compact \textbf{InternViT-300M + MLP + Qwen3 1.7B} architecture, balancing fine-grained visual perception, language-semantic alignment, and deployment cost within an approximately 2B-parameter budget.

To balance business specialization with the retention of general capabilities, we developed a data iteration and curation mechanism and trained the model through a progressive three-stage pipeline: pre-training, mid-training, and post-training. Ablation studies and offline business evaluations show that Xuanwu VL-2B achieves an average score of 67.90 across seven OpenCompass multimodal metrics (vs. 64.27 for InternVL 3.5 2B), an average recall of 94.38\% over seven independent business moderation tasks, and a weighted overall recall of 82.82\% on policy-violating text in challenging adversarial OCR scenarios, outperforming Gemini-2.5-Pro\footnote{Gemini-2.5-Pro was the leading publicly accessible commercial multimodal model available during the development stage of this study and was therefore chosen as a high-standard control model. In this paper it is evaluated through the official API in a zero-shot setting without domain adaptation and with the default dynamic thinking configuration; the comparison is mainly intended to show the gains from domain-specific fine-tuning in the target business scenario.} (76.72\%). These results show that, under a limited parameter budget, Xuanwu VL-2B achieves a practical balance among business alignment, visual perception, general capability retention, and deployment cost.

%% file: sections/01-introduction-en.tex
\section{Introduction}
\label{sec:introduction}


In recent years, general multimodal large language models (MLLMs) such as LLaVA~\cite{liu2023visual}, Qwen-VL~\cite{bai2023qwenvl}, and InternVL~\cite{chen2024internvl, internvl2025internvl35} have demonstrated the strong potential of cross-modal architectures on visual question answering and related tasks. Trained on large-scale image-text corpora and ranging from billions to hundreds of billions of parameters, these models have achieved strong open-domain perception and reasoning performance on public benchmarks.

However, when these general-purpose models are deployed in real-world content moderation and adversarial scenarios, such as filtering heavily distorted diversion images or identifying policy-violating text hidden in AIGC-generated forgeries, they often face clear domain adaptation challenges. This loss of robustness and generalization exposes three core challenges that an industrial-grade content foundation must address:
\begin{itemize} 
    \item \textbf{Inference Cost Bottlenecks and Long-Tail Knowledge Forgetting in Content Understanding}: Very large models bring high inference latency and compute overhead, making it difficult to meet the concurrency demands of large-scale social platforms. Direct domain adaptation can also trigger severe catastrophic forgetting, especially on long-tail knowledge, and weaken the model's original general vision-language representations.
    \item \textbf{Insufficient Granularity in Moderation Knowledge}: To preserve broad open-domain capabilities, mainstream models usually have limited awareness of business-specific context, such as regional regulations, community slang, and fine-grained legal or ethical boundaries. This limits their usefulness in high-precision moderation.
    \item \textbf{Weak Robustness under Adversarial Content}: When confronted with extremely low-resolution text, heavily distorted watermarks, and rapidly evolving AIGC-based adversarial samples, models trained mainly on natural images and public datasets often fail to recover the critical local features, leading to lower recall and unstable judgments.
\end{itemize}

\subsection{Related Work}
General multimodal foundations such as LLaVA~\cite{liu2023visual}, Qwen-VL~\cite{bai2023qwenvl}, and InternVL~\cite{chen2024internvl,internvl2025internvl35} have validated the paradigm of aligning a vision encoder with a language model, achieving strong performance on open-domain visual question answering, perception, and reasoning tasks. Under deployment constraints, MobileVLM~\cite{chu2023mobilevlm} and MiniCPM-V~\cite{yao2024minicpm} further show that lightweight VLMs can remain practically useful through careful model scaling, visual-token design, and training recipes. Meanwhile, SAIL-VL~\cite{sailvl2024} indicates that, even at the 2B scale, pretraining data quality and recipe design remain decisive factors for the final multimodal capability ceiling.

On safety and moderation, prior work has repeatedly shown that multimodal models remain vulnerable to image perturbations, visual inducements, and cross-modal attacks, from visual adversarial jailbreak studies~\cite{qi2023visual} to robustness audits of commercial vision-language systems~\cite{dong2024robustness}. MM-SafetyBench~\cite{liu2023mmsafetybench} and SafeBench~\cite{ying2024safebench} extend this line by systematically evaluating image-driven jailbreaks and harmful-query safety in MLLMs, while Hateful Memes~\cite{kiela2020hatefulmemes} highlights that harmful-content detection often depends on joint image-text semantics rather than either modality alone. However, these studies focus mainly on jailbreak behavior, hateful content, or generic safe-response evaluation, and only partially cover the fine-grained category attribution, covert diversion detection, and OCR-style adversarial variants that are central to industrial moderation.

For fine-grained perception and OCR, OCRBench~\cite{liu2023ocrbench} exposes the systematic weakness of large multimodal models on text-rich images. GOT~\cite{wei2024got} and TextMonkey~\cite{liu2024textmonkey} improve scene-text and document understanding through unified OCR modeling, high-resolution visual encoding, and text-centric training. On the data side, JEST~\cite{evans2024jest} shows that carefully curated joint example selection can significantly improve multimodal learning efficiency. Overall, existing work advances general multimodal capability, efficient deployment, safety evaluation, and OCR perception, but still leaves open the industrial setting we target: simultaneously balancing content understanding, content moderation, and adversarial-content defense under a limited parameter budget and practical deployment constraints. We therefore study this problem through a unified combination of business evaluation, compact architecture, three-stage training, and post-training alignment.

\begin{figure}[!t]
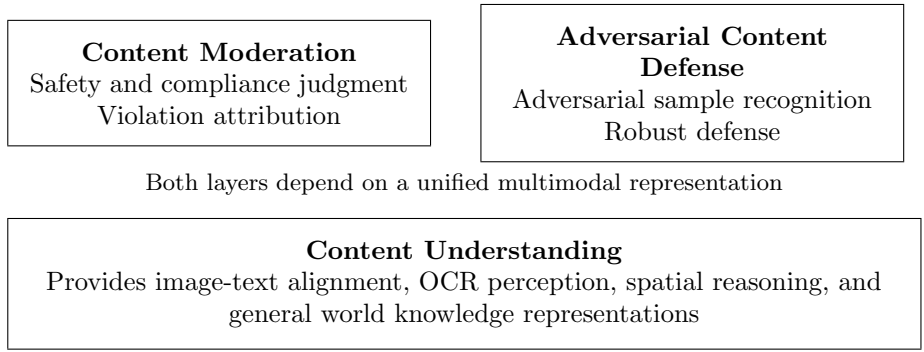

    \centering
    \setlength{\fboxsep}{8pt}
    \renewcommand{\arraystretch}{1.25}
    \begin{tabular}{cc}
        \fbox{\parbox{0.34\textwidth}{\centering \textbf{Content Moderation}\\Safety and compliance judgment\\Violation attribution}} &
        \fbox{\parbox{0.34\textwidth}{\centering \textbf{Adversarial Content Defense}\\Adversarial sample recognition\\Robust defense}} \\[6pt]
        \multicolumn{2}{c}{\small Both layers depend on a unified multimodal representation} \\[6pt]
        \multicolumn{2}{c}{\fbox{\parbox{0.78\textwidth}{\centering \textbf{Content Understanding}\\Provides image-text alignment, OCR perception, spatial reasoning, and general world knowledge representations}}}
    \end{tabular}
    \caption{Capability layering of the Xuanwu industrial multimodal foundation. The lower layer provides unified multimodal representations, while the upper layers address moderation and adversarial defense.}
    \label{fig:ecosystem_structure}
\end{figure}

As illustrated in Figure~\ref{fig:ecosystem_structure}, we abstract the capability architecture of Xuanwu into a layered model. The bottom layer, "Content Understanding," provides unified multimodal representations for downstream applications. Built on top of this foundation, "Content Moderation" and "Adversarial Content Defense" together form the business-facing safety stack.

We divide the capabilities of this industrial-grade foundation into the following three core dimensions:
\begin{itemize}
    \item \textbf{Content Understanding (Cross-Modal Core Engine)}: preserving strong representations of general world knowledge and spatial relationships even after business-oriented specialization.
    \item \textbf{Content Moderation (Security \& Compliance Control)}: going beyond static label classification to support fine-grained judgments grounded in legal, ethical, and platform-specific standards.
    \item \textbf{Adversarial-Content Defense}: handling high-frequency variants, covert diversion tactics, and AIGC-based image forgeries while preserving robust feature extraction under severe interference.
\end{itemize}

The core contributions of this paper can be summarized in the following four points:
\begin{enumerate}
    \item \textbf{A business-oriented evaluation system for content moderation and adversarial OCR}: We build an evaluation system spanning content understanding, content moderation, and adversarial-content defense, enabling a structured assessment of how 2B-scale multimodal models behave in business-domain settings.
    \item \textbf{A three-stage training pipeline}: We propose a \textbf{Pre-Training $\rightarrow$ Mid-Training $\rightarrow$ Post-Training} methodology that combines data refinement, business knowledge injection, and post-training alignment to balance business performance gains with the retention of general capabilities.
    \item \textbf{A compact 2B multimodal architecture}: We adopt the compact \textbf{InternViT-300M + MLP + Qwen3 1.7B} architecture and show that, under an approximately 2B parameter budget, it can balance business moderation, adversarial OCR, and general multimodal capability.
    \item \textbf{Business-oriented data curation and post-training alignment}: Through high-fidelity SFT data construction, structured CoT supervision, and adversarial-OCR-oriented GRPO alignment, we improve violation detection, attribution, and recall in challenging business scenarios.
\end{enumerate}

%% file: sections/02-architecture-en.tex
\section{Model Architecture}

\subsection{Architecture Design}
For multimodal models designed for adversarial scenarios, our primary architectural goal is to prioritize \textbf{training stability, cross-modal alignment efficiency, and deployment efficiency} under limited parameter and compute budgets. Prior work shows that although MoE and complex routing mechanisms can increase model capacity at roughly fixed activated compute, they also introduce routing complexity, communication overhead, and optimization instability. Switch Transformer~\cite{fedus2021switch} simplifies conventional MoE to top-1 routing to mitigate these issues; ST-MoE~\cite{zoph2022stmoe} and StableMoE~\cite{dai2022stablemoe} further show that sparse expert models often suffer from unstable training, uncertain fine-tuning behavior, and routing fluctuation, and may require auxiliary losses or staged training. In multimodal settings, LIMoE~\cite{mustafa2022limoe} and MM1~\cite{mckinzie2024mm1} likewise indicate that stable optimization and balanced expert or modality usage remain key challenges, while MM1's ablations suggest that the vision encoder, image resolution, and visual token count matter more than connector complexity. Recent work such as Sparse Upcycling~\cite{komatsuzaki2023sparseupcycling} and DS-MoE~\cite{pan2024dsmoe} also notes that sparse MoE models often require more total parameters to match dense models, and that training them from scratch remains data- and engineering-intensive. Based on these observations, and considering the practical constraints of a 2B-scale model for high-frequency adversarial moderation, we adopt the design principle of \textbf{"Minimalist Architecture with Stable Components"}. After extensive empirical comparison, we choose a compact recipe: \textbf{InternViT-300M} as the visual backbone, connected to the \textbf{Qwen3 1.7B} language backbone through a lightweight \textbf{MLP projector}. This design provides a practical balance among parameter count, training stability, and deployment cost.

\begin{figure}[!t]
    \centering
    \includegraphics[width=0.8\textwidth]{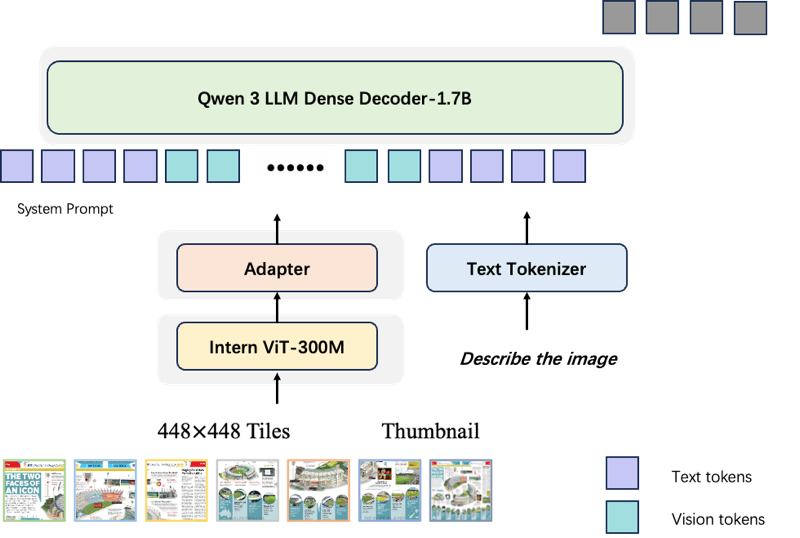}
    \caption{Core architecture of Xuanwu VL-2B: InternViT-300M serves as the visual backbone and Qwen3 1.7B serves as the language backbone, connected by an MLP projector.}
    \label{fig:architecture}
\end{figure}

\subsection{Vision Encoder Exploration and Selection}
Highly deceptive scenarios, such as adversarial content, distorted text, and AIGC-generated forged images, place stringent demands on the \textbf{fine-grained feature extraction capability} of the vision encoder. Traditional CLIP-style models generally emphasize coarse semantic alignment and are often insufficient for these settings. To improve the fine-grained perceptual capability of the vision encoder, we conducted a series of comparative studies around ViT-based backbones.

\begin{itemize}
    \item \textbf{Single Vision Encoder Benchmarking (ViT Benchmarking):} We systematically evaluated several state-of-the-art visual backbones, including \textbf{InternViT-300M}, \textbf{AIMv2-Huge}~\cite{aimv2_2024}, and \textbf{SAIL-ViT-Huge}~\cite{sailvl2024}. These candidates represent different pre-training paradigms and parameter scales. We compared them quantitatively in terms of multimodal performance, text performance, content moderation, adversarial-defense performance, and deployment cost. The detailed results are reported in Section~\ref{sec:vit_selection}, Table~\ref{tab:vit_compare}. Considering both empirical results and inference overhead, we selected \textbf{InternViT-300M} as the mainline vision encoder.
    
    \item \textbf{Exploration of Multi-ViT Fusion (Multi-ViT Exploration):} Motivated by EAGLE~\cite{shi2025eagle}'s systematic study of multi-encoder designs, we further assessed whether a dual-encoder setup could benefit business moderation. Prior work suggests that different vision encoders can provide complementary representations, potentially improving both general understanding and OCR or document perception. GOT~\cite{wei2024got}'s General OCR Theory also highlights the importance of character perception and recognition in text-centric image understanding. Based on these observations, we paired InternViT, which provides strong general multimodal representations, with OCR-oriented GOT-ViT as a second encoder, and evaluated two simple fusion strategies: concatenation along the visual-token dimension and concatenation along the feature-channel dimension while keeping the sequence length unchanged. The quantitative comparison is reported in Section~\ref{sec:ablation_dual_vit}. Considering both empirical results and deployment overhead, the final system retains a single \textbf{InternViT-300M} to keep the training and inference pipeline simple.
\end{itemize}

\subsection{Language Foundation and Cross-Modal Alignment}
On the language side, we selected \textbf{Qwen3 1.7B}~\cite{qwen2025qwen3}. At this scale, it provides strong reasoning and instruction-following ability. We use a lightweight two-layer MLP as the cross-modal connector instead of a more complex structure such as Q-Former, so as to keep the architecture simple while preserving visual features as much as possible.

\subsection{Architecture Hyperparameter Details}
To clearly illustrate the model scale and internal structural characteristics of the Xuanwu VL-2B multimodal foundation, Table~\ref{tab:architecture} details the specific hyperparameter settings of the three core modules composing the system: the vision encoder, the cross-modal projection layer, and the language backbone.

\begin{table}[ht]
\centering
\caption{Detailed Hyperparameter Configurations for Xuanwu VL-2B Core Architecture}
\label{tab:architecture}
\begin{tabular}{lccc}
\hline
\textbf{Configurations}  & \textbf{InternViT-300M} & \textbf{MLP Projector} & \textbf{Qwen3-1.7B} \\ \hline
\textbf{Architecture Type}   & ViT-L/14                & 2-Layer Perceptron     & Transformer Decoder   \\
\textbf{Hidden Size}     & 1024                    & 1024 $\rightarrow$ 2048& 2048                \\
\textbf{Num Heads}       & 16                      & -                      & 16 (Q) / 8 (KV)       \\
\textbf{Num Layers}      & 24                      & 2                      & 28                  \\
\textbf{Parameters}      & $\sim$304M              & $\sim$4M               & $\sim$1.7B          \\ \hline
\end{tabular}
\end{table}

\subsection{Dynamic High-Resolution Perception Mechanism}
In real-world industrial adversarial defense, challenges such as tiny malicious watermarks in image corners and extremely small distorted text in dense layouts continually test the model's ability to capture fine-grained details. Simply resizing every image to a fixed $448 \times 448$ resolution causes severe loss of local visual information.

To address this issue, we introduced a \textbf{Dynamic High-Resolution Perception (Dynamic Tiling \& Resizing) Mechanism} at the visual front end. Instead of applying a rigid partitioning scheme, the mechanism adapts to the original aspect ratio of the image:
\begin{enumerate}
    \item \textbf{Adaptive Tiling}: The system computes a tiling grid that best matches the current image (e.g., $1\times 3$, $2\times 2$) and splits the image into up to 12 local tiles of $448 \times 448$ pixels.
    \item \textbf{Global Thumbnail}: In addition to the local high-resolution tiles, the system preserves a down-sampled global overview image. The full set of inputs ($N$ local tiles $+ 1$ global thumbnail) is then fed into InternViT-300M for feature extraction.
    \item \textbf{Sequence Reshaping (Pixel Unshuffle)}: To prevent the number of visual tokens from growing excessively with larger $N$, we apply Pixel Unshuffle~\cite{shi2016subpixel} to the extracted visual features, reducing the token count to one quarter of the original size. As a result, each $448 \times 448$ tile contributes only 256 visual tokens.
\end{enumerate}
This dynamic perception design helps the model preserve both global context and local detail under a controlled compute budget. It allows the model to first capture the overall scene and then inspect fine-grained cues, from poster-scale layouts to tiny adversarial text.

%% file: sections/03-training-en.tex
\section{Three-Stage Training}

We divide the model lifecycle into three distinct, continuous stages: pre-training, mid-training, and post-training. The overall three-stage, five-step training recipe is summarized in Table~\ref{tab:training_data_overview}.

\begin{table}[ht]
\centering
\footnotesize
\caption{Overview of the Three-Stage, Five-Step Training Recipe}
\label{tab:training_data_overview}
\renewcommand{\arraystretch}{1.15}
\setlength{\tabcolsep}{3pt}
\begin{tabular}{p{1.7cm}p{1.8cm}p{2.1cm}p{3.8cm}p{1.2cm}p{2.6cm}}
\hline
\textbf{Stage} & \textbf{Sub-stage} & \textbf{Trainable Modules} & \textbf{Data Type} & \textbf{Scale} & \textbf{Optimization Config} \\ \hline
Pre-Training & Modality Alignment & MLP projector & High-quality caption alignment data & 1.3M & LR = $1 \times 10^{-3}$, Batch = 256 \\
Pre-Training & General Knowledge Learning & Full Model & General image-text paired data & 17.33M & LR = $1 \times 10^{-3}$, Batch = 256 \\ \hline
Mid-Training & Business Knowledge Learning & Full Model & Base-retention + instruction-following enhancement + moderation + SPAM adversarial data & 2.8M & LR = $2 \times 10^{-5}$, Batch = 128 \\ \hline
Post-Training & Instruction Tuning & Full Model & High-fidelity SFT data & 8.408M & LR = $1 \times 10^{-5}$ \\
Post-Training & RL Alignment & Full Model & GRPO alignment data / adversarial OCR alignment samples & 810k & LR = $1 \times 10^{-6}$ \\ \hline
\end{tabular}
\end{table}

Unless otherwise specified, all data scales in this section refer to the effective samples that actually participate in training after filtering, deduplication, and quality control. The appendix data inventory tables instead report the raw source-data scale before entering the training pipeline, so their totals are larger.

\subsection{Pre-Training}
The primary objective of this stage is to endow the model with foundational image-text understanding and alignment capabilities. The pre-training pipeline is further divided into two sub-stages. First, \textbf{MLP Projector Alignment} uses approximately 1.3 million high-quality caption samples, freezing the ViT and LLM backbones while training only the MLP projector to establish initial cross-modal alignment. This is followed by \textbf{Full-parameter Joint Training}, which uses approximately 17.33 million image-text pairs and jointly optimizes the ViT, MLP, and LLM end to end. Under the effective-training accounting used in the main text, these two sub-stages together use about 18.63 million multimodal samples; the larger appendix count corresponds to the raw source-data inventory before pipeline filtering. This builds a strong foundation for visual dialogue, image captioning, and visual question answering (VQA).
Both sub-stages share a base learning rate of $1 \times 10^{-3}$ and a global batch size of 256, while the pre-training optimization uses cosine decay with 100 warmup steps.

\textbf{Loss Function Design:}
This stage employs the standard multimodal autoregressive Next-token Prediction cross-entropy loss, designed to enable the language foundation to understand visual features and generate fluent outputs:
\begin{equation}
L_{pretrain} = -\sum_{i=1}^{N} \log P_\theta(y_i | y_{<i}, X_v)
\label{eq:pretrain}
\end{equation}
where $X_v$ represents the visual input features, $y_i$ is the $i$-th token of the target text sequence, $y_{<i}$ denotes the generated text tokens preceding the $i$-th token, $P_\theta$ is the network parameter set to be optimized, $N$ is the total sequence length, and $L_{pretrain}$ is the pre-training loss for this stage.

\subsection{Mid-Training}
Following foundational pre-training, the model is further trained on approximately 2.8 million samples during the mid-training stage. These data consist of four parts: 1.75M base-retention samples obtained by sampling 10\% of the pre-training corpus to preserve general representations; 148k instruction-following enhancement samples built from AutoIF-Instruct~\cite{pretrain_autoif}, IFEval-like Data~\cite{zhou2023instruction}, and Tulu-3 Personas~\cite{pretrain_tulu3} to improve complex instruction understanding and output-format compliance; 646k content moderation samples, including positive and negative moderation data together with OCR and caption annotations for both image and text scenarios; and 257k SPAM adversarial samples, including synthetic SPAM adversarial data, human-annotated SPAM adversarial data, and LLM-assisted SPAM adversarial data. All LLM-assisted annotations are manually reviewed. A detailed breakdown is provided in Appendix Table~\ref{tab:appendix_midtraining_data}. To ensure quality and distributional balance across these heterogeneous sources, we built an automated data refinement pipeline:
\begin{itemize}
    \item \textbf{Quality Filtering}: Utilizing the LLM Judge alongside CLIP scoring mechanisms to eliminate low-quality, noisy, and contradictory samples.
    \item \textbf{Diversity Control}: Employing the K-Means clustering algorithm to ensure a distributional balance between business data and general data, helping mitigate "catastrophic forgetting" during the model's business specialization process.
\end{itemize}

This data mixture preserves general capabilities while further strengthening instruction following, content moderation, and adversarial SPAM recognition.

\textbf{Loss Function Design:}
Similar to the first stage, the mid-training stage continues to use the autoregressive Cross-Entropy Loss, but the sampling weights for high-quality business data are increased to sharpen the model's sensitivity to complex business image-text structures.

\subsection{Post-Training}
During this stage, the model is aligned from a general-purpose multimodal model into a business-oriented moderation model that better follows moderation rules and human values in complex adversarial environments.
The overall post-training data composition and the four-category breakdown of the general-data block are provided in Appendix Tables~\ref{tab:appendix_posttraining_overview} and \ref{tab:appendix_posttraining_general}.

\subsubsection{Supervised Fine-Tuning}
In the SFT stage, we started from approximately 8 million raw samples under a setting where full manual annotation would have been prohibitively expensive. We therefore built an automated data-cleaning pipeline centered on a model-in-the-loop workflow. Strong teacher models such as Qwen2.5-VL-7B~\cite{qwen25vl2025} and InternVL3-14B~\cite{zhu2025internvl3} were used for cross-model verification and loss rescoring, enabling us to filter the corpus down to a substantially higher-purity instruction set; for example, a core set of about 70,000 high-quality samples was extracted from an 840,000-sample base corpus. The SFT stage uses a learning rate of $1 \times 10^{-5}$, and the resulting instruction data teach the model explicit moderation rules, output formats, and the task pattern of ``observe first, reason next, conclude finally.''

\subsubsection{Reinforcement Learning}
We compared GRPO~\cite{shao2024deepseekmath}, DPO~\cite{rafailov2023direct}, and GSPO~\cite{zheng2025gspo}. In the SPAM classification task, \textbf{GRPO (Group Relative Policy Optimization)} delivered the clearest gains. The reinforcement-learning stage uses a learning rate of $1 \times 10^{-6}$, with the KL-divergence penalty coefficient $\beta$ set to $0.01$. Unlike standard maximum-likelihood training, we design a composite reward for adversarial interception scenarios that combines classification reward, format reward, and OCR reward.

\textbf{Loss Function and Reward Design:}
In the GRPO (Group Relative Policy Optimization) phase, we optimize the model with the standard GRPO objective. For each sampled output $y_i$, the total reward is an equally weighted sum of classification consistency, format compliance, and OCR alignment:
\begin{equation}
r_i = R_{\mathrm{cls}}(y_i, x) + R_{\mathrm{fmt}}(y_i) + R_{\mathrm{ocr}}(y_i, x)
\label{eq:reward}
\end{equation}
Here, $R_{\mathrm{cls}}$ rewards correct spam classification labels, $R_{\mathrm{fmt}}$ rewards compliance with the predefined structured output format, and $R_{\mathrm{ocr}}$ measures character-level alignment between the generated result and the target violating text. Specifically, the OCR reward uses a bag-similarity objective based on character-set matching:
\begin{equation}
R_{\mathrm{ocr}}(y_i, x) = 1 - \frac{N_{\mathrm{halluc}, i} + N_{\mathrm{miss}, i}}{\max(N_{\mathrm{pred}, i}, N_{\mathrm{gt}})}
\label{eq:reward_ocr}
\end{equation}
Here, $N_{\mathrm{halluc}, i}$ is the number of hallucinated characters in the $i$-th output, $N_{\mathrm{miss}, i}$ is the number of ground-truth violating characters missed by the model, $N_{\mathrm{pred}, i}$ is the number of predicted characters in the $i$-th output, and $N_{\mathrm{gt}}$ is the number of violating characters in the ground truth. The remaining optimization details follow the standard GRPO setup, where the group-relative advantage is obtained by normalizing rewards within each sampled group. To stay aligned with the main-text adversarial OCR metric, all GRPO-related summary results are reported as weighted overall recall across the eight adversarial subsets. Under this definition, GRPO improves the adversarial OCR result from 76.42\% to 82.82\%.

\subsubsection{Explainability and Chain-of-Thought}
In complex business security scenarios, a black-box model that only outputs a final interception label is often difficult for stakeholders to trust and analyze. We therefore introduced a large amount of structured Chain-of-Thought (CoT) data during both SFT and RL. Instead of directly predicting a final category, Xuanwu VL-2B follows a three-stage reasoning pattern: ``observe first, reason next, conclude finally.'' For example, when the model is given a highly obfuscated diversion image, it first locates the tiny distorted text, then explains why the content violates moderation rules, and finally outputs the violation label and category. In real deployments, this design both improves performance on logically complex violations and reduces the review burden of manual re-checking and appeals by providing traceable reasoning.

\subsection{Training Setup}
\begin{itemize}
    \item \textbf{Hardware Cluster}: Full-scale training was run on a distributed cluster of 64 NVIDIA A100 80GB GPUs (8 nodes with 8 GPUs each). For a model at the 2B scale, the throughput is about 6.1k tokens per second per GPU, and the total training cost is approximately 3,500 GPU hours.
    \item \textbf{Global Parameters}: The training framework uses DeepSpeed~\cite{rasley2020deepspeed}, mixed precision (AMP / bf16), and Flash Attention-2~\cite{dao2023flashattention2}. The packed context length is up to 16,384 tokens.
\end{itemize}

%% file: sections/04-evaluation-en.tex
\section{Evaluation Framework and Business Experiments}

\subsection{Industrial-Grade Evaluation Standards}
For real-world social-platform scenarios, we constructed a specialized evaluation framework covering three dimensions: content moderation, adversarial-content defense, and content understanding:
\begin{itemize}
    \item \textbf{Content Moderation Evaluation}: We construct a content moderation evaluation set from manually reviewed samples, containing 5,881 instances across seven moderation labels: ad (973), high-risk (845), illegal (872), porn (802), vulgar (907), other (589), and normal (893). The normal category represents non-violating content, while the remaining categories correspond to the major risk types in content moderation. During evaluation, we formulate an independent binary decision task for each category and report recall for that category; average-7 denotes the arithmetic mean over the seven category recalls. To avoid exposing the original moderation prompts and sensitive samples in the main text, we describe the tasks using ``category definition + structured template.'' Taking ad as an example, given an input sample $x$, the model outputs a binary label $y \in \{\text{Yes}, \text{No}\}$ according to a predefined ad classification criterion. A sample is labeled as ad when its main content serves commercial promotion, recruitment, transaction posting, or product or service marketing; if such information appears only as background or as an auxiliary element rather than the main focus, it is not counted as ad.
    \item \textbf{Adversarial Content Evaluation}: By mimicking both human-crafted and algorithmically generated variant attacks (e.g., added noise and geometric distortion), we built an evaluation set across eight adversarial dimensions (2,504 images in total: 191 AIGC fused images, 98 combination layouts, 791 handwriting samples, 60 long images, 357 noise-interference samples, 380 small-text samples, 243 warped-text samples, and 384 heavily watermarked images). We define the \textbf{OCR recall} metric for each subset as:
    \begin{equation}
    RecallRate = \frac{N_{recognized}}{N_{total}}
    \label{eq:recall}
    \end{equation}
    where $N_{recognized}$ is the number of violating characters successfully recognized by the model, and $N_{total}$ is the total number of violating characters in that adversarial subset. The per-category rows in the table report subset-level OCR recall, while the main-text summary results report weighted overall recall across the eight adversarial subsets.
    \item \textbf{Content Understanding Evaluation}: We evaluate the retention of general capabilities after business specialization using public benchmarks. The multimodal benchmark suite includes HallusionBench~\cite{guan2023hallusionbench}, AI2D~\cite{kembhavi2016diagram}, MMStar~\cite{chen2024mmstar}, OCRBench~\cite{liu2023ocrbench}, MMBench~\cite{liu2023mmbench}, MMMU~\cite{yue2023mmmu}, and MathVista~\cite{lu2023mathvista}; the text-only suite includes IFEval~\cite{zhou2023instruction}, GPQA Diamond~\cite{rein2023gpqa}, GSM8K~\cite{cobbe2021gsm8k}, MMLU~\cite{hendrycks2020mmlu}, C-Eval~\cite{huang2023ceval}, BBH~\cite{suzgun2022bbh}, MATH~\cite{hendrycks2021math}, HumanEval~\cite{chen2021humaneval}, and MBPP~\cite{austin2021mbpp}. The main text and stage-wise comparison consistently report multimodal average-7 and text average-9.
\end{itemize}

We note that the 148k instruction-following enhancement data used during training are not the IFEval benchmark itself, but are instead composed of AutoIF-Instruct~\cite{pretrain_autoif}, IFEval-like Data~\cite{zhou2023instruction}, and Tulu-3 Personas~\cite{pretrain_tulu3}. For public benchmark data that also appear in training, we only use the official GSM8K train split and the official C-Eval dev split for data construction; all reported evaluation results are measured on the corresponding official evaluation splits that were not used for training.

\subsection{General Benchmarks}
We comprehensively evaluated Xuanwu VL-2B on mainstream multimodal and text-only benchmarks. The results show that even after substantial business specialization, Xuanwu VL-2B remains competitive on both multimodal and text-only evaluations, with several results close to or better than open-source models at a comparable scale.

\begin{table}[ht]
\centering
\caption{Comparison of General Multimodal Capabilities (\%)}
\label{tab:general_multimodal}
\begin{tabular}{lccc}
\hline
\textbf{Benchmark} & \textbf{InternVL 3.0 2B} & \textbf{InternVL 3.5 2B} & \textbf{Xuanwu VL-2B} \\ \hline
HallusionBench  & 43.32                          & 46.78                             & \textbf{47.32}              \\
AI2D            & 77.75                          & 77.95                             & \textbf{82.19}              \\
MMStar          & 58.73                          & 56.20                             & \textbf{60.47}              \\
OCRBench        & 81.80                          & 83.10                             & \textbf{89.80}              \\
MMBench v1.1    & \textbf{79.41}                 & 75.08                             & 79.02                       \\
MMMU (val)      & 45.89                          & \textbf{50.51}                    & 48.11                       \\
MathVista       & 52.80                          & 60.30                             & \textbf{68.40}              \\ \hline
\textbf{average-7}& 62.81                        & 64.27                             & \textbf{67.90}              \\ \hline
\end{tabular}
\end{table}

As shown in Table~\ref{tab:general_multimodal}, serving as a stress test for general foundation capabilities, we evaluated the model against mainstream multimodal capability assessment datasets. Xuanwu VL-2B not only maintained its scores across the vast majority of capabilities but also improved the average score across the seven metrics by 3.63 points compared with InternVL 3.5 2B.

\begin{table}[ht]
\centering
\caption{Comparison of Text-Only Capabilities (\%)}
\label{tab:general_text}
\begin{tabular}{lccc}
\hline
\textbf{Benchmark} & \textbf{InternVL 3.0 2B} & \textbf{InternVL 3.5 2B} & \textbf{Xuanwu VL-2B} \\ \hline
IFEval           & 49.07                          & 69.88                             & \textbf{76.35}              \\
GPQA Diamond    & 29.80                          & \textbf{34.34}                    & 32.83                       \\
GSM8K           & 73.09                          & \textbf{74.07}                    & 73.92                       \\
MMLU            & 61.74                          & \textbf{65.51}                    & 62.94                       \\
C-Eval          & \textbf{73.12}                 & 63.66                             & 62.18                       \\
BBH             & 51.89                          & \textbf{62.85}                    & 59.68                       \\
MATH            & 42.06                          & \textbf{43.98}                    & 41.10                       \\
HumanEval       & 62.80                          & \textbf{67.07}                    & 65.85                       \\
MBPP            & 45.60                          & 49.80                             & \textbf{50.60}              \\ \hline
\textbf{average-9}& 54.35                        & \textbf{59.02}                    & 58.38                       \\ \hline
\end{tabular}
\end{table}

As shown in Table~\ref{tab:general_text}, while incorporating image-text alignment, Xuanwu VL-2B keeps its overall average-9 score close to InternVL 3.5 2B (58.38 vs. 59.02) and gains 6.47 points on instruction-following tasks such as IFEval. The largest declines appear on C-Eval~\cite{huang2023ceval} (-10.94), BBH~\cite{suzgun2022bbh} (-3.17), and several math- or code-related tasks, including MMLU~\cite{hendrycks2020mmlu}, MATH~\cite{hendrycks2021math}, HumanEval~\cite{chen2021humaneval}, and MBPP~\cite{austin2021mbpp}. We attribute this mainly to two factors: (1) the large amount of business-domain data introduced during mid-training partially compresses general academic knowledge; and (2) about 8\% of outputs do not strictly follow the required answer format (e.g., ``ANSWER: A''), which causes extraction failures under hard matching. Future iterations will address this through additional instruction-following data and stronger format supervision.

\subsection{Content Moderation and Adversarial Evaluation}

\subsubsection{Content Moderation}
Under real business scenarios, Xuanwu VL-2B shows stable advantages on content moderation, especially on long-tail high-risk categories.

\begin{table}[ht]
\centering
\caption{Business Content Moderation Recall for Independent Binary Decisions (\%)}
\label{tab:business_moderation}
\begin{tabular}{lccc}
\hline
\textbf{Category}       & \textbf{InternVL 3.0 2B} & \textbf{InternVL 3.5 2B} & \textbf{Xuanwu VL-2B} \\ \hline
ad                      & 70.61                            & 88.39                               & \textbf{99.38}                \\
high-risk               & 4.50                             & 7.93                                & \textbf{97.99}                \\
illegal                 & 1.38                             & 12.93                               & \textbf{92.19}                \\
porn                    & 40.80                            & 61.80                               & \textbf{97.15}                \\
vulgar                  & 21.63                            & 25.72                               & \textbf{79.08}                \\
other                   & 4.09                             & 47.50                               & \textbf{94.89}                \\
normal                  & 66.42                            & 91.60                               & \textbf{99.95}                \\ \hline
\textbf{average-7}      & 29.92                            & 47.98                               & \textbf{94.38}                \\ \hline
\end{tabular}
\end{table}

As shown in Table~\ref{tab:business_moderation}, the content moderation evaluation set constructed from manually reviewed samples covers seven categories: ad, high-risk, illegal, porn, vulgar, other, and normal. Manual review shows that under independently constructed binary decision tasks for each category, Xuanwu VL-2B improves recall across all dimensions; specifically, its average-7 score is 46.40 percentage points higher than InternVL 3.5 2B.

\subsubsection{Adversarial OCR Evaluation}
For adversarial-content evaluation, we focus on OCR recall over transformed policy-violating text and report both subset-level recall across the eight attack types and the corresponding weighted overall recall. To compare against both open-source baselines and a commercial frontier model in a single view, we include InternVL 3.0 2B, InternVL 3.5 2B, Gemini-2.5-Pro, and Xuanwu VL-2B on the same test set. During evaluation, the open-source models were deployed with lmdeploy, temperature set to 0 (greedy decoding), and a maximum output length of 8,192 tokens; the Gemini-2.5-Pro control group was evaluated through the official API in a zero-shot setting without additional domain adaptation, using the default dynamic thinking configuration. We note that Gemini-2.5-Pro was also used for machine rewriting and auxiliary annotation in parts of the business-data construction pipeline, so this comparison should be interpreted as a task-specialized comparison after teacher-assisted data construction rather than as a claim of universal superiority.

\begin{table}[ht]
\centering
\small
\setlength{\tabcolsep}{4pt}
\caption{Business Adversarial OCR Recall Comparison (\%)}
\label{tab:business_spam_baseline}
\begin{tabular}{lcccc}
\hline
\textbf{Category} & \textbf{InternVL 3.0 2B} & \textbf{InternVL 3.5 2B} & \textbf{Gemini-2.5-Pro} & \textbf{Xuanwu VL-2B} \\ \hline
aigc                & 16.21                            & 14.34                               & 26.06                      & \textbf{32.83}              \\
combination         & 60.60                            & 58.05                               & 74.31                      & \textbf{74.62}              \\
handwriting         & 80.99                            & 80.17                               & \textbf{89.53}             & 88.86                       \\
long                & 59.87                            & 57.28                               & \textbf{80.15}             & 79.22                       \\
noise               & 58.77                            & 55.14                               & 73.27                      & \textbf{79.09}              \\
small               & 69.99                            & 77.29                               & \textbf{92.82}             & 91.76                       \\
warp                & 59.17                            & 58.98                               & 75.27                      & \textbf{81.81}              \\
watermark           & 61.10                            & 61.35                               & 63.77                      & \textbf{93.13}              \\
\hline
\textbf{weighted overall}  & 64.74                            & 64.79                               & 76.72                      & \textbf{82.82}              \\ \hline
\end{tabular}
\end{table}

As shown in Table~\ref{tab:business_spam_baseline}, the adversarial-content test set is partitioned into eight difficult dimensions: AIGC fused images, combination layouts, handwriting, long images, noise interference, small text, warped text, and heavy watermarks. Xuanwu VL-2B reaches a weighted overall recall of 82.82\%, clearly above InternVL 3.0 2B and InternVL 3.5 2B at 64.74\% and 64.79\%, and also above Gemini-2.5-Pro at 76.72\%. Looking at individual subsets, Xuanwu VL-2B shows larger gains on aigc, noise, warp, and watermark, while Gemini-2.5-Pro is only slightly ahead on handwriting, long, and small. We note that Gemini-2.5-Pro is used here as a zero-shot commercial control model without domain adaptation; this comparison is therefore intended mainly to illustrate the gains from domain-specific fine-tuning in the target business scenario, rather than to serve as a strictly matched comparison under identical training conditions.

\subsection{Pre-Training and Mid-Training Comparison}
To analyze the effects of pre-training and mid-training more concretely, we compare three checkpoints: the general pre-training checkpoint of InternVL 3.5 2B (denoted as InternVL 3.5 2B PT), the Xuanwu pre-training checkpoint (Xuanwu PT), and the Xuanwu Mid-Training checkpoint. This is not intended as a fair fine-tuning baseline under the same training pipeline; rather, it is a stage-wise comparison meant to show how the capability profile evolves from general pre-training to Xuanwu pre-training and then to mid-training. The Xuanwu Mid-Training checkpoint uses about 2.8M samples, including base-retention data sampled at 10\% from the pre-training corpus, instruction-following enhancement data, content moderation data, and adversarial-content data; the detailed breakdown is provided in Appendix Table~\ref{tab:appendix_midtraining_data}.

\begin{table}[ht]
\centering
\small
\setlength{\tabcolsep}{4pt}
\caption{Stage-wise comparison: InternVL 3.5 2B PT, Xuanwu PT, and Xuanwu Mid-Training (\%)}
\label{tab:stage_compare}
\begin{tabular}{p{4.2cm}ccc}
\hline
\textbf{Metric} & \textbf{InternVL 3.5 2B PT} & \textbf{Xuanwu PT} & \textbf{Xuanwu Mid-Training} \\ \hline
Multimodal average-7          & 58.17 & 62.29 & 62.63 \\
Text average-9                & 58.27 & 51.96 & 56.14 \\
Content Moderation average-7  & 30.04 & 57.11 & 94.44 \\
Adversarial OCR weighted overall & 64.62 & 63.75 & 75.35 \\ \hline
\end{tabular}
\end{table}

As shown in Table~\ref{tab:stage_compare}, compared with InternVL 3.5 2B PT, Xuanwu PT improves multimodal average-7 by 4.12 points and content moderation average-7 by 27.07 points, while text average-9 drops by 6.31 points and adversarial OCR weighted overall recall stays nearly unchanged (63.75 vs. 64.62). This suggests that under the unified core architecture and the Xuanwu pre-training pipeline, the model already gains stronger multimodal and moderation-related capabilities, but at some cost to text performance.

Further, when moving from Xuanwu PT to Xuanwu Mid-Training, multimodal average-7 increases by another 0.34 points, text average-9 recovers by 4.18 points, content moderation average-7 rises by 37.33 points, and adversarial OCR weighted overall recall increases by 11.60 points. Looking at the sub-metrics, IFEval improves from 46.69 to 70.50, MMLU from 57.10 to 63.59, and HumanEval from 62.20 to 62.80, while the multimodal metrics remain largely stable except for a more noticeable gain on MathVista (53.40 to 57.10). This indicates that Xuanwu pre-training mainly establishes the multimodal and moderation foundation, whereas Mid-Training further injects business knowledge and partially restores some of the text capability that declined during pre-training. Full per-metric scores are provided in Tables~\ref{tab:appendix_stagewise_multimodal}, \ref{tab:appendix_stagewise_text}, \ref{tab:appendix_stagewise_moderation}, and \ref{tab:appendix_stagewise_spam}.

\subsection{Qualitative Results}
Purely quantitative metrics are not sufficient to fully characterize the model's behavior in authentic adversarial settings. To more intuitively illustrate Xuanwu VL-2B's explainable moderation capability based on Chain of Thought (CoT), we provide several anonymized online interception cases in Appendix~\ref{sec:appendix_qualitative}. These cases cover AIGC deepfakes, severe distortions, and multi-layer watermarks, together with the model's reasoning and attribution process.

\subsection{Selection and Ablation Studies}

\subsubsection{Visual Encoder Selection}
\label{sec:vit_selection}
To verify the impact of the visual feature extractor on final metrics, we extracted model checkpoints after the mid-training stage for four core capability tests. As shown in Table~\ref{tab:vit_compare}:

\begin{table}[ht]
\centering
\small
\setlength{\tabcolsep}{4pt}
\caption{Vision encoder selection comparison after mid-training. Multimodal = average-7; text = average-9; content moderation and adversarial OCR are reported as recall (\%).}
\label{tab:vit_compare}
\begin{tabular}{lcccc}
\hline
    \textbf{ViT Type} & \textbf{Multimodal} & \textbf{Text} & \textbf{Content Moderation} & \textbf{Adversarial OCR} \\ \hline
InternViT-300M        & \textbf{62.63} & 56.14 & 94.44 & 75.35 \\
AIMv2-Huge (600M)     & 61.88 & 54.87 & \textbf{96.18} & 74.91 \\
SAILViT-Huge (600M)   & 61.68 & 56.25 & 95.35 & \textbf{76.73} \\ \hline
\end{tabular}
\end{table}

Table~\ref{tab:vit_compare} shows that InternViT-300M achieves the highest multimodal score (62.63) while remaining competitive on content moderation and adversarial OCR. AIMv2-Huge is 1.74 points higher on content moderation, but it does not surpass InternViT-300M on multimodal performance and shows a more visible drop on text performance. SAILViT-Huge is slightly better on text and adversarial OCR by 0.11 and 1.38 points, respectively, but the inference overhead of a 600M vision encoder is not justified by these limited gains. Considering general capability, business metrics, and deployment cost together, InternViT-300M provides the best overall trade-off.

\subsubsection{Dual-ViT Fusion}
\label{sec:ablation_dual_vit}
To control experimental cost while staying consistent with EAGLE's base recipe, we adopted its public setup for this structural exploration: pre-alignment on LLaVA-595K~\cite{liu2023visual}, followed by multimodal SFT on Eagle1.8M~\cite{shi2025eagle}. We compare three variants: a single-encoder InternViT baseline, visual-sequence concatenation fusion, and feature-channel concatenation fusion. The first uses only InternViT. The visual-sequence fusion variant concatenates the outputs of two visual encoders along the visual-token dimension, thereby doubling the visual sequence length. The feature-channel fusion variant keeps the visual sequence length unchanged and concatenates the two streams along the feature dimension. The second encoder is GOT-ViT~\cite{wei2024got}, whose training objective is more oriented toward OCR perception and fine-grained text modeling.

\begin{table}[ht]
\centering
\small
\caption{Ablation results for dual-encoder visual fusion. Sequence fusion concatenates the two streams along the visual-token dimension; channel fusion keeps the token count unchanged and concatenates along the feature dimension.}
\label{tab:dual_vit_ablation}
\begin{tabular}{llccc}
\hline
\textbf{Stage} & \textbf{Metric} & \textbf{Single} & \textbf{Seq. Fusion} & \textbf{Channel Fusion} \\
\hline
Pretrain & HallusionBench & 17.84 & 18.28 & 17.43 \\
Pretrain & AI2D & 36.27 & 38.60 & 35.75 \\
Pretrain & MMStar & 31.47 & 30.87 & 29.50 \\
Pretrain & OCRBench & 32.50 & 32.70 & 34.50 \\
\hline
Pretrain & average-4 & 29.52 & 30.11 & 29.30 \\
\hline
SFT & HallusionBench & 21.48 & 26.13 & 25.04 \\
SFT & AI2D & 61.46 & 69.17 & 70.01 \\
SFT & MMStar & 36.53 & 37.40 & 37.30 \\
SFT & OCRBench & 48.00 & 45.10 & 43.10 \\
\hline
SFT & average-4 & 41.87 & 44.45 & 43.86 \\
\hline
\end{tabular}
\end{table}

As shown in Table~\ref{tab:dual_vit_ablation}, both dual-encoder variants improve several understanding-oriented metrics, including HallusionBench, AI2D, and MMStar, indicating that introducing an OCR-oriented encoder does change the representation profile of the model. However, on OCRBench, which is the most critical metric for our target scenario, both dual-encoder variants underperform the single-encoder baseline and fail to deliver the expected OCR gains. Since the dual-encoder design also increases inference overhead, the results suggest that complementary visual representations have not yet translated into better OCR performance under the current fusion and alignment setup. We therefore retain single-encoder InternViT as the mainline visual backbone, which offers a better trade-off between OCR performance and deployment efficiency.

\subsubsection{Core Strategy Ablation}
To verify the technical effectiveness of each core strategy through quantitative experiments, we systematically conducted ablation studies around the adversarial evaluation set. To stay aligned with the main-text adversarial OCR metric, all ablation results in this section are reported as weighted overall recall over the eight adversarial subsets. The specific quantitative outcomes are delineated in Table~\ref{tab:ablation}:

\begin{table}[ht]
\centering
\caption{Ablation Studies on Adversarial-Set}
\label{tab:ablation}
\begin{tabular}{lcc}
\hline
\textbf{Model Configuration} & \textbf{Weighted Overall (\%)} & \textbf{False Positive Rate (FP, \%)} \\ \hline
\textbf{Xuanwu VL-2B (Full Pipeline)}  & \textbf{82.82}       & \textbf{0.08}                  \\
w/o GRPO                          & 76.42       & 0.12                  \\
w/o SFT Data Curation                  & 79.35       & 0.31                  \\ \hline
\end{tabular}
\end{table}

\begin{itemize}
    \item \textbf{Reinforcement Alignment Gains ($\Delta$ GRPO)}: As shown in Table~\ref{tab:ablation}, removing GRPO on the SPAM adversarial cohort causes the weighted overall recall to drop from 82.82\% to 76.42\% (a decrease of 6.40 points), validating the crucial role of RL in improving the model's performance on difficult samples.
    \item \textbf{SFT Data Curation Gains}: After removing SFT data curation strategies such as low-pass filtering augmentation and cross-model validation loss rescanning, the model's False Positive rate on the Adversarial-Set rises from 0.08\% to 0.31\%, while weighted overall recall drops to 79.35\%. This shows that a high signal-to-noise data curation pipeline has a direct impact on system stability.
\end{itemize}

%% file: sections/05-conclusion-en.tex
\section{Conclusion and Future Work}

This paper presents the architecture, training pipeline, and evaluation results of Xuanwu VL-2B. For industrial content moderation and adversarial OCR scenarios, we adopt a compact InternViT + Qwen architecture and combine business-oriented data curation with SFT and GRPO-based alignment to build a multimodal foundation model suited to real-world deployment. Experimental results show that the model achieves strong performance on business moderation and adversarial OCR while maintaining reasonable general capability and deployment cost within an approximately 2B-parameter budget.

\subsection{Future Evolution}
Xuanwu VL-2B will not remain limited to a single-purpose recognition model. Instead, we plan to evolve it toward a more general and agile agent system. We will focus on the following four directions:

\begin{itemize}
    \item \textbf{Native Multimodal Integration}: Move from an add-on multimodal design to a native multimodal architecture. By representing images and text within a unified token sequence, we aim to reduce information loss during modality conversion and improve perception of subtle visual cues in difficult adversarial cases.
    \item \textbf{Full-Sensory Perception}: Extend beyond static images by introducing \textbf{Audio} and \textbf{Video} modalities. With temporal modeling and audio semantic extraction, we aim to build stronger understanding for complex scenarios such as livestreams and short videos.
    \item \textbf{Global \& Multilingual Expansion}: Strengthen multilingual capability to support global deployment, with particular attention to low-resource languages and localized dialects so that the foundation can operate more consistently across regions and cultures.
    \item \textbf{Quantization \& Inference Optimization}: While preserving interception quality, continue to explore lower-bit deployment schemes such as INT4. Through operator fusion, memory optimization, and bottleneck analysis, we aim to reduce the compute cost and VRAM footprint of each inference and improve deployment efficiency at scale.
\end{itemize}

The long-term goal of Xuanwu VL-2B is to evolve from a task-specific multimodal system into a continuously evolving foundation for content ecosystems, with adaptability further strengthened through online policy distillation and self-adversarial evolution.

\subsection{Limitations and Potential Risks}
Although Xuanwu VL-2B is stable in most adversarial scenarios, it still has inherent limitations as a probabilistic generative model:
\begin{itemize}
    \item \textbf{Missed Detections in Extreme Edge Cases}: Under extreme adversarial tactics such as ultra-dense overlapping watermarks or pixel-level hidden text that is nearly invisible to the human eye, the model may still occasionally miss violations because of the physical limits of the $448 \times 448$ local receptive field and Pixel Unshuffle feature granularity. A typical example is a complex product image where multiple semi-transparent watermarks overlap with very small diversion text, leaving only a few pixels for the target characters.
    \item \textbf{Hallucinations and Logical Shortcuts}: In some long-context moderation chains of thought, the base model may occasionally make logical jumps and produce plausible but incorrect violation attributions under specific long-tail rules. For example, when the main image is benign but the border contains several weakly related marketing phrases or disguised aliases, the model may overemphasize local high-risk cues and underweight the full context, leading to category errors.
\end{itemize}

%% file: sections/07-appendix-en.tex
\section{Appendix}

\subsection{Data Processing Pipeline}
\label{sec:appendix_data_processing}

To ensure the scale, quality, and diversity of the Xuanwu model's pre-training and fine-tuning data, we built the core data processing pipeline mainly on top of the DataJuicer framework~\cite{chen2023datajuicer} and deployed it at scale. The workflow can be summarized in the following eight stages:

\begin{enumerate}
    \item \textbf{Stage 1: Data Collection \& Format Unification} \\
    Raw materials are collected from open-source repositories such as Infinity-MM and LAION-5B, together with large-scale internal multimodal data, including sampled daily feed streams and backfilled business-report data. We convert all data into a standardized JSONL format under the WebDataset mechanism for distributed reading, using \texttt{<image>} or \texttt{<video>} tags for multimodal placeholder alignment.
    
    \item \textbf{Stage 2: Data Filtering} \\
    We first perform coarse filtering with a combination of heuristic rules and model assistance. This includes filtering by image or video aspect ratio and short-edge resolution, for example removing heavily blank or truncated images with severely imbalanced width-to-height ratios. For specific dynamic scenarios, a safety model is used to intercept NSFW and other violating content; long-term internal experiments show that about 17.2\% of high-risk dynamic data can be removed at this stage. On the text side, we use a KenLM model to compute perplexity and remove illogical token sequences, garbled text, and low-quality titles.
    
    \item \textbf{Stage 3: Data Deduplication} \\
    To improve information density in the large-scale corpus, we perform strict cross-source deduplication. In addition to rule-based URL deduplication, the text modality uses MinHash with Locality-Sensitive Hashing (LSH) to remove semantic overlap. The image modality computes perceptual hashes (PHash) at scale and compares Hamming distances for cleaning. For example, in a single deduplication stream, about 530,000 newly added samples are first deduplicated internally and then compared against an existing 150,000-sample baseline training set and online evaluation sets to eliminate potential test-set leakage.
    
    \item \textbf{Stage 4: Clustering \& Sampling} \\
    To address complex multimodal distribution bias, we use high-quality BGE embeddings for text and models such as Qwen-VL to extract global visual representations. After merging these features, we perform dimensionality reduction and cluster the large-scale dataset with hundreds of thousands of K-Means centers. To avoid collapse on long-tail data, we apply balanced downsampling to extremely frequent head clusters, such as white-background product images or selfie portraits.
    
    \item \textbf{Stage 5: LLM API Data Labeling} \\
    To obtain high-quality instruction-tuning pairs, we perform large-scale machine rewriting and labeling through APIs of commercial closed-source models such as Gemini-2.5-Pro. We design detailed system-prompt constraints, often exceeding 800 tokens in a single context, to elicit detailed Chain-of-Thought reasoning with at least 200 tokens of analysis. Among hundreds of thousands of raw samples processed by this pipeline, only about 70,000 high-quality text-only and multimodal instructions are retained after strict selection and reconstruction.
    
    \item \textbf{Stage 6: JudgeModel Scoring \& Filtering} \\
    We introduce high-performance cross-modal matching models, such as large CLIP variants, to measure strong image-text correlation and remove low-correlation or clearly hallucinated noisy texts. For image features with dense text distributions and high-risk interception tags, we additionally apply critic-style discriminative scoring. For some images, we also apply low-pass filtering during preprocessing, removing 1.5\%$\sim$7\% of high-frequency noise features so that the model focuses more on contour structure rather than AIGC-induced corner noise.
    
    \item \textbf{Stage 7: Difficulty Grading} \\
    We build a dual-track model framework, for example combining a 2B model with a 30B expert model, to run a loss-based rescoring pipeline. By comparing inference confidence profiles and cross-entropy loss across the same data batch, we dynamically estimate a theoretical difficulty score for each sample. Samples are then divided into three levels, easy, medium, and hard, which naturally support curriculum learning for later complex tasks.
    
    \item \textbf{Stage 8: Daily Evaluation Monitoring} \\
    We maintain a highly automated pipeline-dashboard loop for daily monitoring. Every batch of cleaned, deduplicated, and labeled tuning data is fed into an online sampling-based validation channel that tracks multiple metrics simultaneously. When the model degrades on a particular class distribution, or when new high-frequency violating attack patterns cause local data drift, the dashboard raises an alarm and triggers targeted collection of new adversarial data for the next iteration cycle.
\end{enumerate}

\input{sections/07-appendix-pretrain-en}

\subsection{Mid-Training Data Composition}
\label{sec:appendix_midtraining_data}

To supplement the data-mixture description of the mid-training stage in Section 3, Table~\ref{tab:appendix_midtraining_data} provides a detailed breakdown of the approximately 2.8M training samples used at this stage. To avoid product-specific internal jargon, all data sources are described here by their functional roles.

\begin{table}[ht]
\centering
\scriptsize
\caption{Training data composition of the Mid-Training stage}
\label{tab:appendix_midtraining_data}
\renewcommand{\arraystretch}{1.12}
\setlength{\tabcolsep}{4pt}
\begin{tabular}{p{1.7cm}p{1.8cm}p{1.2cm}p{8.0cm}}
\hline
\textbf{Training Stage} & \textbf{Data Block} & \textbf{Scale} & \textbf{Description} \\ \hline
Mid-Training & General data & 1.898M & 1.75M samples are sampled at 10\% from the pretraining corpus to preserve the model's general representations. Another 148k instruction-following enhancement samples, built from AutoIF-Instruct~\cite{pretrain_autoif}, IFEval-like Data~\cite{zhou2023instruction}, and Tulu-3 Personas~\cite{pretrain_tulu3}, are used to improve complex instruction understanding and output-format compliance. \\
Mid-Training & Content moderation & 646k & Positive and negative image/text moderation data contribute 598k samples, including 418k image samples and 180k text samples, covering both positive and negative moderation cases. Another 48k OCR and caption annotation samples are used to supplement OCR and caption capability in dynamic-image scenarios. \\
Mid-Training & Content adversarial & 257k & Synthetic SPAM adversarial data contribute 50k samples, primarily built from watermark-distorted OCR images. Human-annotated SPAM adversarial data contribute 40k samples for adversarial recognition. LLM-assisted SPAM adversarial data contribute 167k samples and are all manually reviewed. \\ \hline
\end{tabular}
\end{table}

\noindent\textit{Note:} The three data blocks sum to approximately 2.801M samples.

\subsection{Post-Training Data Composition}
\label{sec:appendix_posttraining_data}

To supplement the data-mixture description of the post-training stage in Section 3, Tables~\ref{tab:appendix_posttraining_overview} and \ref{tab:appendix_posttraining_general} summarize the overall post-training data composition and the four major categories inside the general-data block. The overall post-training corpus contains approximately 9.218M samples, including about 8.408M SFT samples and 810k RL samples. General data account for 8.01M samples and form the core of post-training; they are further organized into Text \& General Instruction, Mathematics, OCR \& Document Understanding, and Advanced Multimodal Capability.

\begin{table}[ht]
\centering
\scriptsize
\caption{Overall composition of post-training data}
\label{tab:appendix_posttraining_overview}
\renewcommand{\arraystretch}{1.12}
\setlength{\tabcolsep}{3pt}
\begin{tabular}{@{}p{0.9cm}p{1.9cm}p{1.0cm}p{8.8cm}@{}}
\hline
\textbf{Training Stage} & \textbf{Data Block} & \textbf{Scale} & \textbf{Description} \\ \hline
SFT & General data & 8.01M & Composed of 5.14M multimodal samples and 2.87M text-only samples, used to build the foundation for general dialogue, instruction following, OCR/document understanding, and multimodal reasoning. \\
SFT & Content moderation & 359k & Content-safety and moderation-classification data, used to strengthen content safety, moderation classification, and policy consistency. \\
SFT & Content adversarial & 39k & Includes caption, full-image OCR, and violating-text OCR data, used to improve OCR perception and violating-text recognition in real business scenarios. \\
RL & Content adversarial & 810k & Composed of spam-classification and OCR violating-text alignment data, used to further improve task accuracy and robustness in business settings. \\ \hline
\end{tabular}
\end{table}

\begin{table}[ht]
\centering
\scriptsize
\caption{Four major categories of the general-data block and representative references}
\label{tab:appendix_posttraining_general}
\renewcommand{\arraystretch}{1.12}
\setlength{\tabcolsep}{3pt}
\begin{tabular}{@{}>{\raggedright\arraybackslash}p{2.7cm}>{\raggedleft\arraybackslash}p{1.5cm}>{\raggedright\arraybackslash}p{9.9cm}@{}}
\hline
\textbf{Category} & \textbf{Count} & \textbf{Datasets and References} \\ \hline
Text \& General Instruction & 2,303,232 & InfinityInstruct \cite{pretrain_infinity_instruct}; AutoIF-Instruct \cite{pretrain_autoif}; IFEval-like Data \cite{zhou2023instruction}; Tulu-3 Personas \cite{pretrain_tulu3}; C-Eval (dev split) \cite{huang2023ceval} \\
Mathematics & 54,888 & GSM8K (train split) \cite{cobbe2021gsm8k}; Orca-Math \cite{pretrain_orca_math} \\
OCR \& Document Understanding & 494,072 & ChartQA \cite{pretrain_chartqa}; DocVQA \cite{pretrain_docvqa}; InfoVQA \cite{pretrain_infovqa}; TextVQA \cite{pretrain_textvqa}; OCR-VQA \cite{pretrain_ocr_vqa}; SROIE \cite{pretrain_sroie}; POIE \cite{pretrain_poie}; FUNSD \cite{pretrain_funsd}; Imgur5k \cite{pretrain_imgur5k_dataset}; IAM \cite{pretrain_iam}; HME100K \cite{pretrain_hme100k}; Sujet-Finance-QA \cite{pretrain_sujet_finance_qa_vision_100k}; LAION-COCO \cite{pretrain_relaion_coco}; DVQA \cite{pretrain_dvqa}; ST-VQA \cite{pretrain_st_vqa} \\
Advanced Multimodal Capability & 4,645,928 & LLaVA-OneVision \cite{pretrain_llava_onevision}; Cambrian-10M \cite{pretrain_cambrian1}; Infinity-MM \cite{pretrain_infinity_mm}; MMInstruct-GPT4V \cite{pretrain_mminstruct_gpt4v}; InternVL2.5 SFT \cite{pretrain_internvl25}; Koniq-10k \cite{pretrain_koniq_10k}; A-OKVQA \cite{pretrain_aokvqa}; M3\_Cot \cite{pretrain_m3cot} \\ \hline
\end{tabular}
\end{table}

\clearpage
\subsection{Detailed Stage-wise Scores}
\label{sec:appendix_stagewise}

For easier cross-checking of the stage-wise comparison in Section 4, this subsection provides the full per-metric scores for three checkpoints: the general pre-training checkpoint of InternVL 3.5 2B (abbreviated as InternVL PT in the table headers below), the Xuanwu pre-training checkpoint (Xuanwu PT), and the Xuanwu Mid-Training checkpoint. The content moderation metric is recall under category-wise independent binary decisions. For adversarial OCR, each category row reports subset-level OCR recall on policy-violating text, and the final summary row reports weighted overall recall across the eight adversarial subsets. All values are reported in percentages. To stay aligned with the main-text reporting convention, the multimodal table consistently reports the seven multimodal benchmarks and average-7.

\begin{table}[ht]
\centering
\small
\caption{Detailed multimodal scores for the stage-wise comparison (\%)}
\label{tab:appendix_stagewise_multimodal}
\begin{tabular}{lccc}
\hline
\textbf{Metric} & \textbf{InternVL PT} & \textbf{Xuanwu PT} & \textbf{Xuanwu Mid-Training} \\ \hline
HallusionBench               & 41.66 & 47.46 & 46.92 \\
AI2D                         & 72.02 & 78.17 & 77.66 \\
MMStar                       & 47.93 & 54.73 & 54.53 \\
OCRBench                     & 78.10 & 82.90 & 83.50 \\
MMBench v1.1                 & 70.82 & 74.46 & 74.69 \\
MMMU (val)                   & 45.44 & 44.89 & 44.00 \\
MathVista                    & 51.20 & 53.40 & 57.10 \\
\hline
average-7                    & 58.17 & 62.29 & 62.63 \\ \hline
\end{tabular}
\end{table}

\begin{table}[ht]
\centering
\small
\caption{Detailed text-only scores for the stage-wise comparison (\%)}
\label{tab:appendix_stagewise_text}
\begin{tabular}{lccc}
\hline
\textbf{Metric} & \textbf{InternVL PT} & \textbf{Xuanwu PT} & \textbf{Xuanwu Mid-Training} \\ \hline
IFEval       & 60.31 & 46.69 & 70.50 \\
GPQA Diamond & 34.34 & 21.72 & 24.75 \\
GSM8K        & 76.12 & 77.41 & 78.92 \\
MMLU         & 66.77 & 57.10 & 63.59 \\
C-Eval       & 66.81 & 60.43 & 60.57 \\
BBH          & 59.13 & 58.15 & 58.73 \\
MATH         & 58.50 & 36.54 & 36.24 \\
HumanEval    & 52.44 & 62.20 & 62.80 \\
MBPP         & 50.00 & 47.40 & 49.20 \\
\hline
average-9    & 58.27 & 51.96 & 56.14 \\ \hline
\end{tabular}
\end{table}

\begin{table}[ht]
\centering
\small
\caption{Detailed business moderation recall for independent binary decisions in the stage-wise comparison (\%)}
\label{tab:appendix_stagewise_moderation}
\begin{tabular}{lccc}
\hline
\textbf{Category} & \textbf{InternVL PT} & \textbf{Xuanwu PT} & \textbf{Xuanwu Mid-Training} \\ \hline
ad        & 50.05 & 92.29 & 98.56 \\
high-risk & 7.81  & 40.95 & 97.51 \\
illegal   & 8.60  & 29.53 & 98.36 \\
porn      & 13.03 & 82.29 & 98.92 \\
vulgar    & 10.54 & 38.23 & 72.31 \\
other     & 21.96 & 17.16 & 95.91 \\
normal    & 98.28 & 99.35 & 99.52 \\
\hline
average-7 & 30.04 & 57.11 & 94.44 \\ \hline
\end{tabular}
\end{table}

\begin{table}[ht]
\centering
\small
\caption{Detailed business adversarial OCR scores for the stage-wise comparison (\%)}
\label{tab:appendix_stagewise_spam}
\begin{tabular}{lccc}
\hline
\textbf{Category} & \textbf{InternVL PT} & \textbf{Xuanwu PT} & \textbf{Xuanwu Mid-Training} \\ \hline
aigc         & 15.57 & 22.69 & 21.66 \\
combination  & 54.36 & 63.47 & 73.45 \\
handwriting & 80.26 & 71.81 & 88.50 \\
long         & 52.43 & 57.93 & 68.65 \\
noise        & 54.14 & 57.75 & 69.92 \\
small        & 79.36 & 77.48 & 91.39 \\
warp         & 55.31 & 57.56 & 74.05 \\
watermark    & 62.40 & 64.42 & 66.47 \\
\hline
weighted overall    & 64.62 & 63.75 & 75.35 \\ \hline
\end{tabular}
\end{table}

\clearpage
\subsection{Qualitative Results}
\label{sec:appendix_qualitative}
\begin{CJK*}{UTF8}{gbsn}

This section presents several anonymized interception cases of Xuanwu VL-2B in real-world industrial attack-and-defense settings. By introducing Chain of Thought (CoT), the model outputs not only classification labels but also the corresponding moderation rationale. These adversarial OCR cases follow the structured output format: \textbf{[Observation]} describes the main subjects and background of the image $\rightarrow$ \textbf{[Extraction]} identifies all visible or obscure text and symbols in the image $\rightarrow$ \textbf{[Reasoning]} compares the extracted evidence against moderation standards $\rightarrow$ \textbf{[Conclusion]} outputs the final decision (Safe / Violating-Category).

\subsubsection{Handwriting Variant Attack}
\begin{figure}[htbp]
    \centering
    \includegraphics[width=0.9\textwidth, height=0.45\textheight, keepaspectratio]{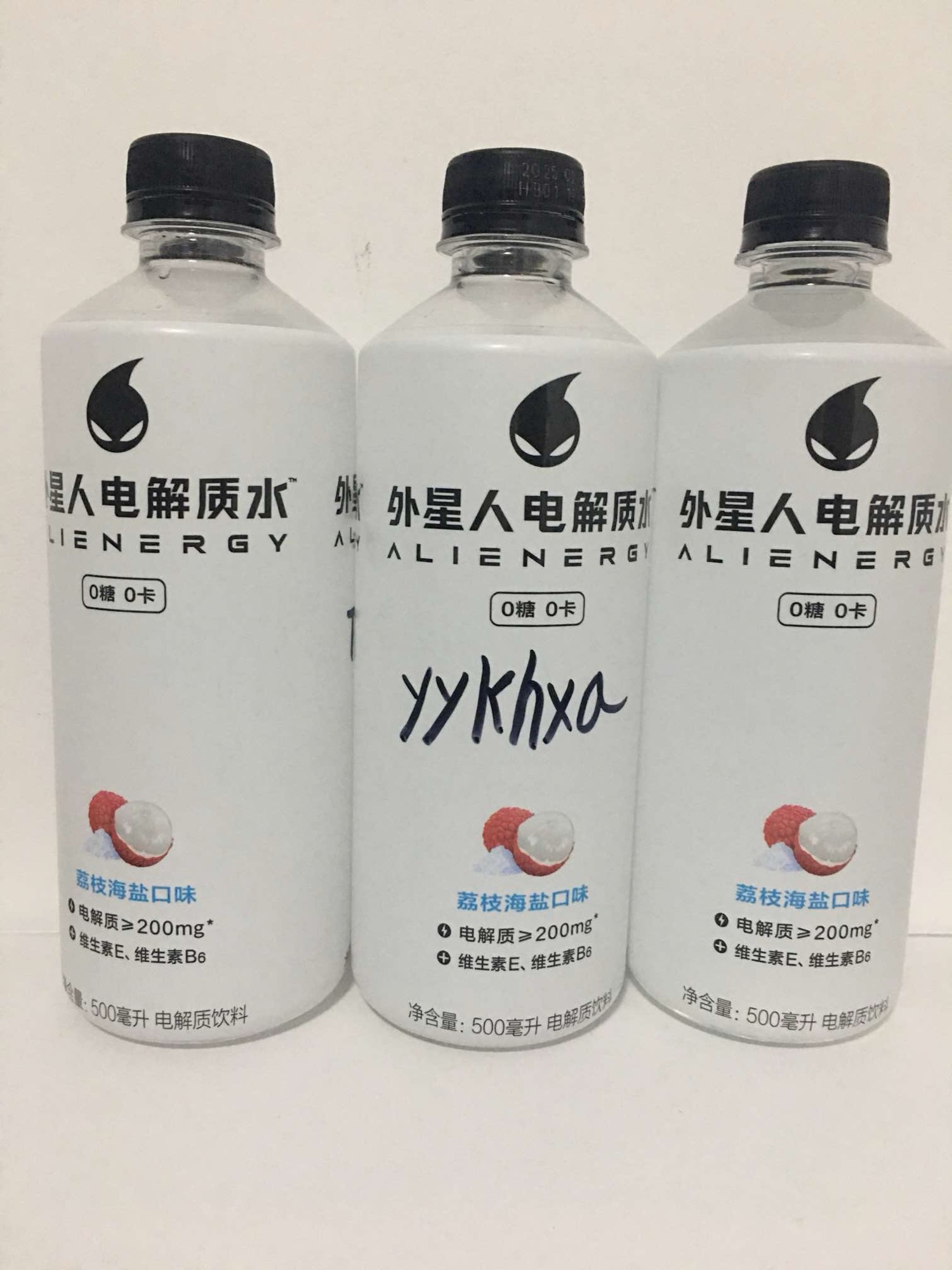}
    \caption{Handwriting-variant case: cursive diversion codes hidden in a complex product image}
    \label{fig:case_handwriting}
\end{figure}
\begin{quote}
\textbf{[Observation]} The image is a promotional product poster for "Alien Electrolyte Water" and contains many standard printed ingredient descriptions and parameter annotations.\\
\textbf{[Extraction]} Amid the dense printed text, the model extracts an anomalous handwritten string that does not match the overall layout: "yyKhxa".\\
\textbf{[Reasoning]} The violator overlays handwriting that is difficult to segment precisely onto a complex product-packaging background. Its irregular cursive strokes and shape variations are intended to bypass conventional OCR detection, and this tactic is often used to hide diversion codes or slang.\\
\textbf{[Conclusion]} Determination: \textbf{Violation (off-platform diversion)}.
\end{quote}

\subsubsection{Layered Watermark Attack}
\begin{figure}[htbp]
    \centering
    \includegraphics[width=0.8\textwidth, height=0.35\textheight, keepaspectratio]{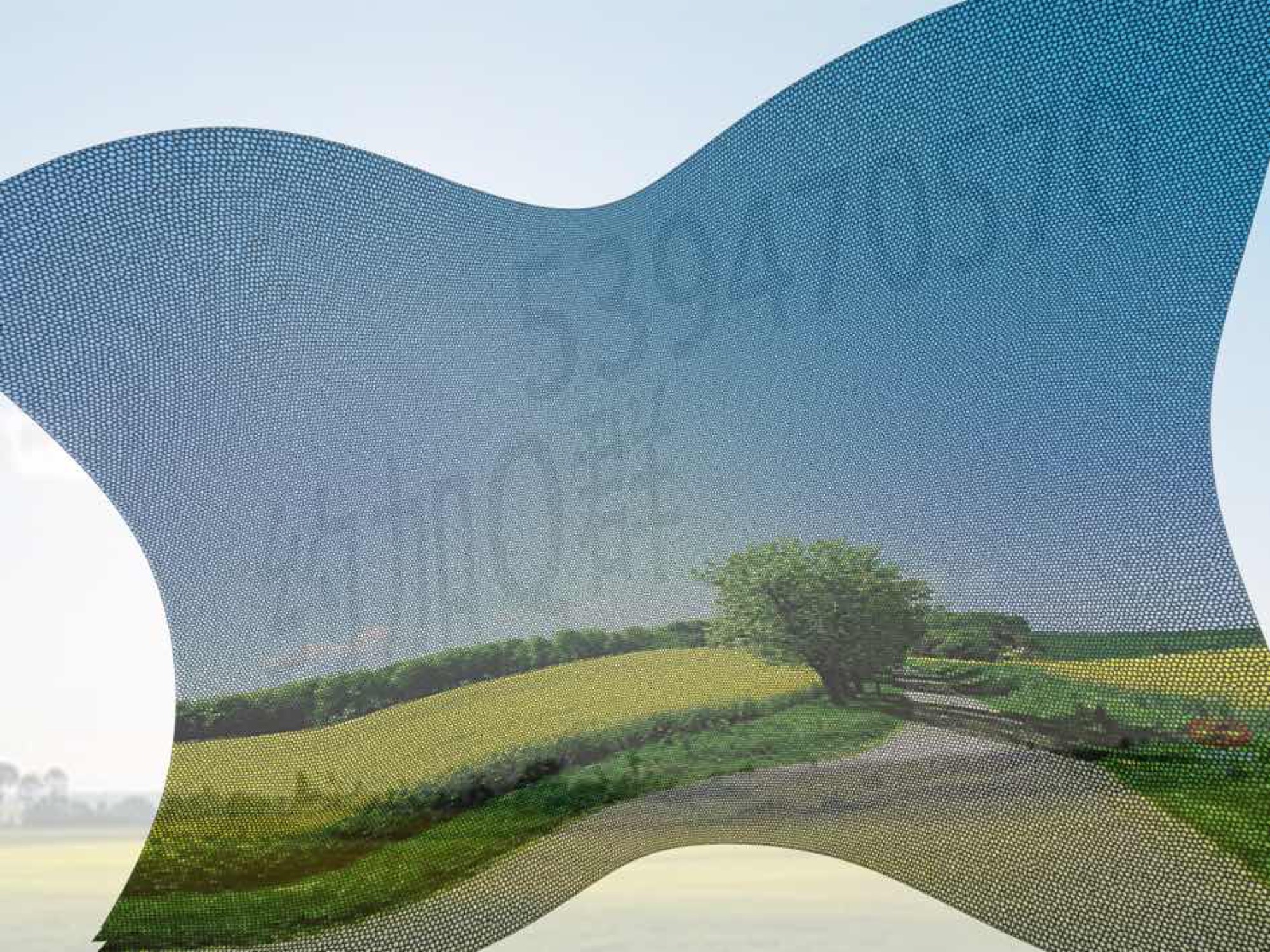}
    \caption{Layered-watermark case: low-opacity contact information hidden in the image}
    \label{fig:case_watermark}
\end{figure}
\begin{quote}
\textbf{[Observation]} The image appears ordinary overall, but the edges or specific regions are covered by semi-transparent information.\\
\textbf{[Extraction]} From the faint transparent text texture, the model restores the watermark string "539470570 约加Q群" (roughly, "join the QQ group").\\
\textbf{[Reasoning]} This is a typical layered-watermark attack. By sharply reducing text opacity and weakening contrast against the background, the attacker attempts to make moderation-system preprocessing miss the signal while covertly spreading third-party social-group contact information.\\
\textbf{[Conclusion]} Determination: \textbf{Violation (off-platform diversion)}.
\end{quote}

\subsubsection{Micro-Font Attack}
\begin{figure}[H]
    \centering
    \includegraphics[width=0.8\textwidth, height=0.35\textheight, keepaspectratio]{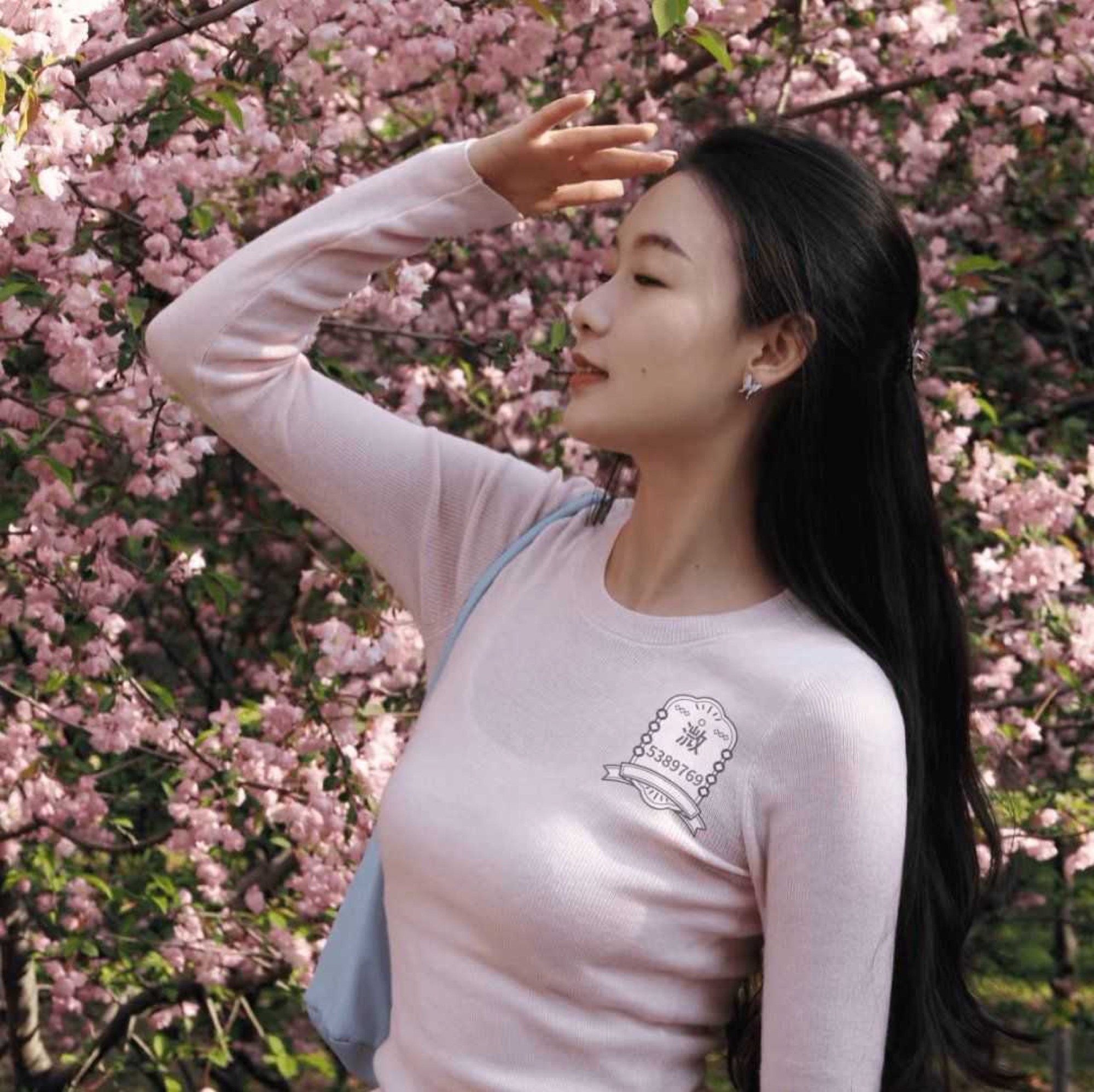}
    \caption{Micro-font case: miniature variant-character diversion}
    \label{fig:case_small}
\end{figure}
\begin{quote}
\textbf{[Observation]} The image shows a broad, otherwise ordinary background scene.\\
\textbf{[Extraction]} In an inconspicuous corner of the image, the model extracts the tiny text "溦 5389769", where "溦" is a commonly used variant character standing in for "微" in "微信" (WeChat).\\
\textbf{[Reasoning]} This case uses a micro-font attack. The core diversion information, including a variant reference to WeChat and an account number, is shrunk to an extreme scale so that compression during transmission or downsampling during model inference will erase the relevant features and defeat normal character detection and localization.\\
\textbf{[Conclusion]} Determination: \textbf{Violation (off-platform diversion)}.
\end{quote}

\subsubsection{Text-Noise Camouflage Attack}
\begin{figure}[htbp]
    \centering
    \includegraphics[width=0.95\textwidth, height=0.45\textheight, keepaspectratio]{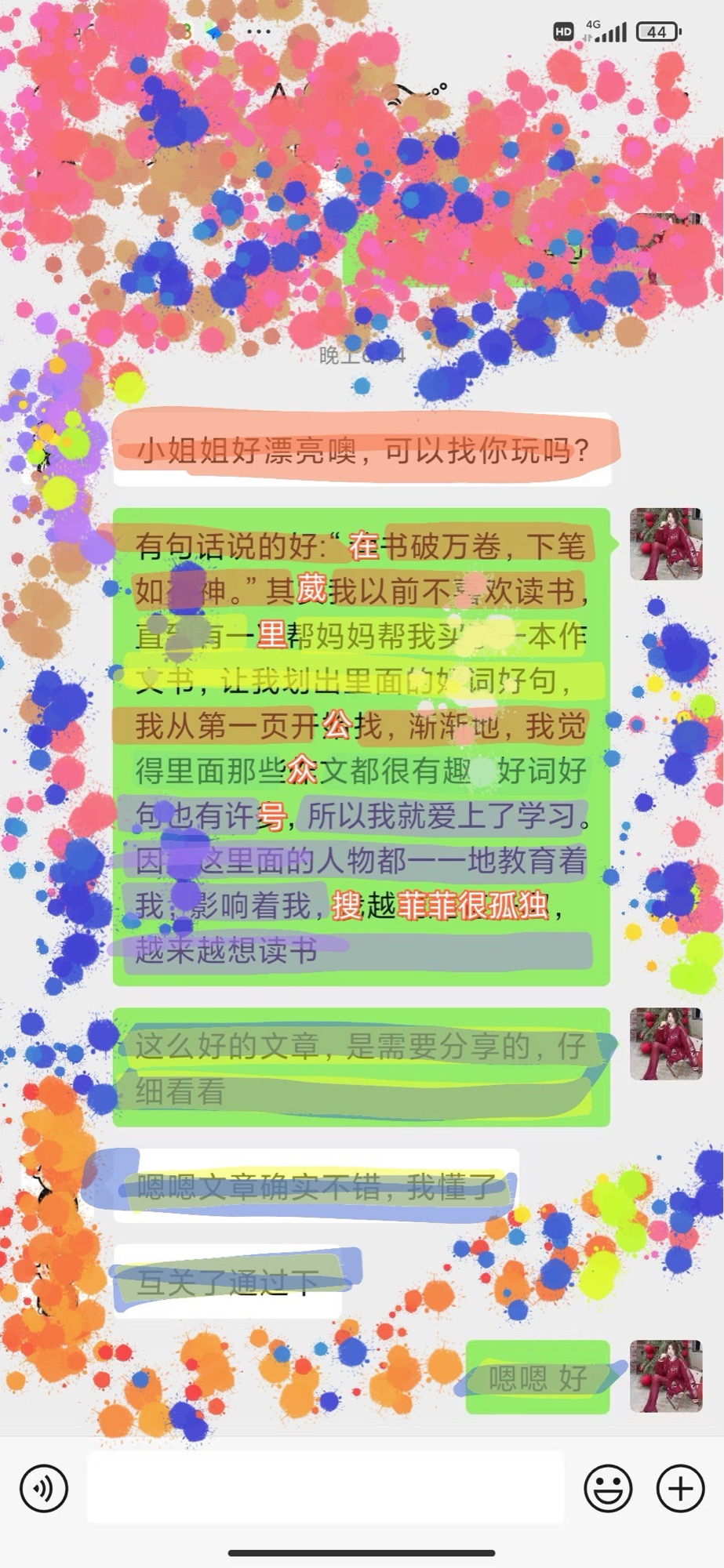}
    \caption{Text-noise camouflage case: diversion keywords hidden inside long prose paragraphs}
    \label{fig:case_noise}
\end{figure}
\begin{quote}
\textbf{[Observation]} The image is filled with long, essay-like paragraphs of text.\\
\textbf{[Extraction]} From the large amount of irrelevant background text, the model extracts the high-risk span "在 葳 里 公众号 搜 菲菲很孤独", where "葳" is a variant character used to allude to "微" in WeChat-related diversion cues.\\
\textbf{[Reasoning]} This is a text-noise camouflage attack. The violator deliberately constructs hundreds of characters of irrelevant normal text as ``noise cover'' to dilute the frequency of sensitive keyword hits. Within this long passage, disguised social-platform aliases and diversion account information are inserted to interfere with attention-based detection.\\
\textbf{[Conclusion]} Determination: \textbf{Violation (off-platform diversion)}.
\end{quote}

\subsubsection{Severe Distortion and Warping Attack}
\begin{figure}[htbp]
    \centering
    \includegraphics[width=0.8\textwidth, height=0.35\textheight, keepaspectratio]{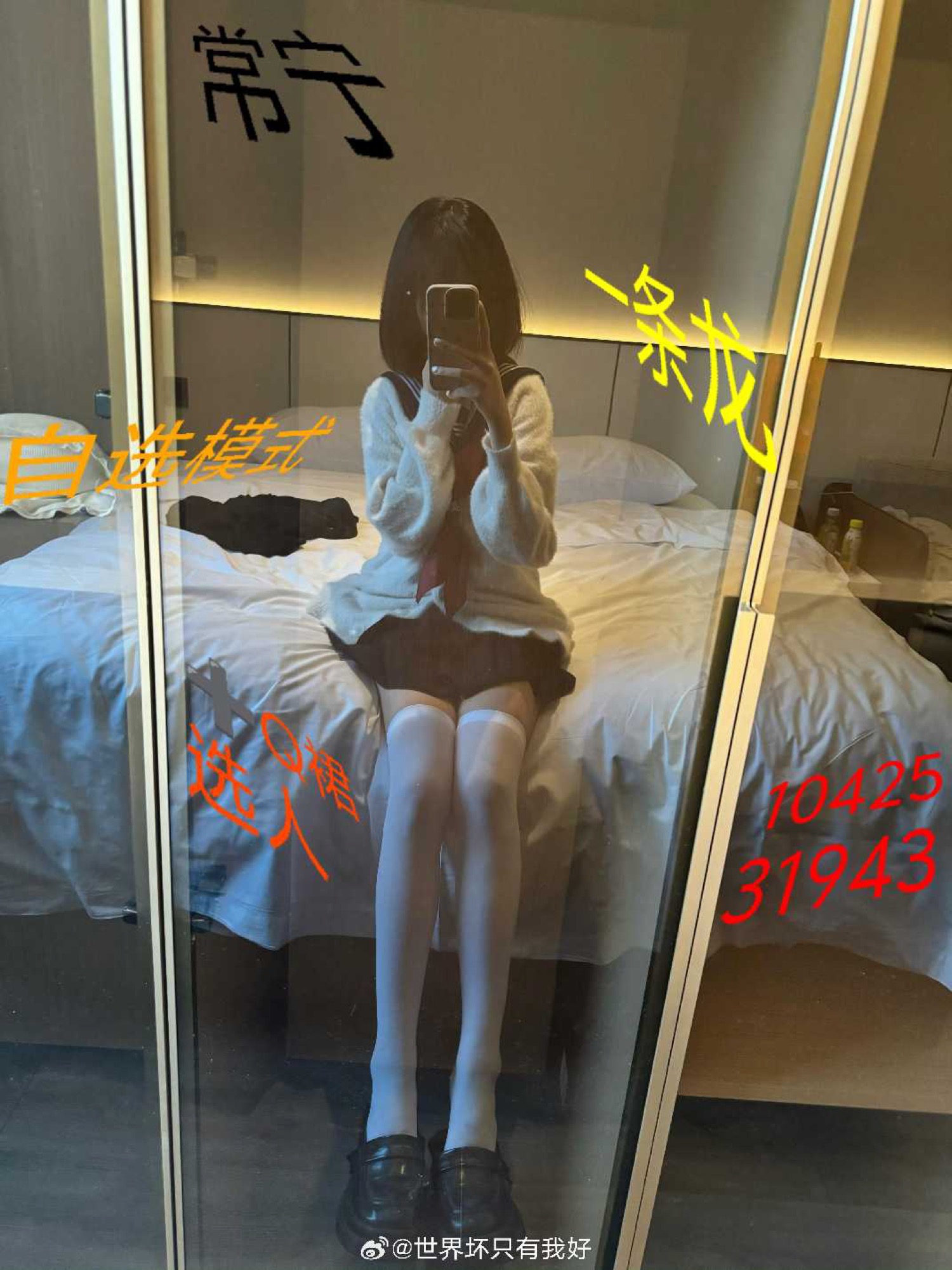}
    \caption{Severe distortion and warping case: breaking character structure to evade detection}
    \label{fig:case_warp}
\end{figure}
\begin{quote}
\textbf{[Observation]} The image contains a highly cluttered advertisement overlaid with illicit slang.\\
\textbf{[Extraction]} Despite strong distortion and slanted warping, the model correctly recognizes a combination of cues, including the strings "一条龙", "Q裙", and "选人", together with the number strings "10425" and "31943".\\
\textbf{[Reasoning]} This case uses a severe distortion and warping attack. The attacker aggressively destroys the original geometric structure of the character strokes. Some of these strings are typical slang for pornographic services or solicitation, while "Q裙" and the following digits point to a QQ group used for pornographic diversion. The goal is to make conventional font-template matching fail.\\
\textbf{[Conclusion]} Determination: \textbf{Violation (off-platform diversion)}.
\end{quote}

\subsubsection{AIGC Deepfake Attack}
\begin{figure}[htbp]
    \centering
    \includegraphics[width=0.8\textwidth, height=0.35\textheight, keepaspectratio]{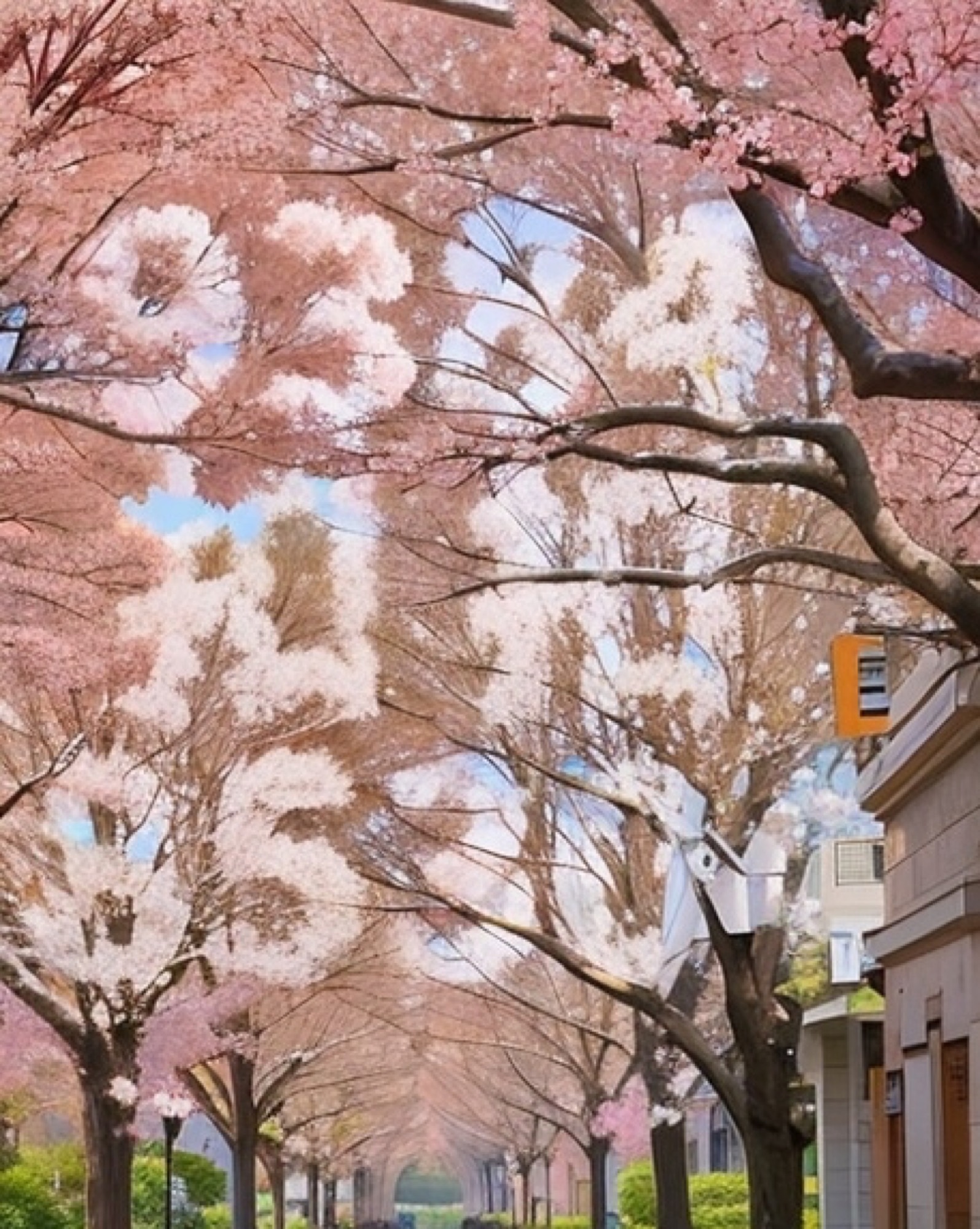}
    \caption{AIGC-forgery case: contact information fused into generated image textures}
    \label{fig:case_aigc}
\end{figure}
\begin{quote}
\textbf{[Observation]} The image presents a carefully blended combination of a concrete scene and abstract AIGC-generated features.\\
\textbf{[Extraction]} The model decodes the concealed string "9384 扣扣 85304" from the natural lines and very high-frequency generated details of the image, where "扣扣" is a common obfuscated reference to QQ.\\
\textbf{[Reasoning]} This is an AIGC deepfake case. Malicious users employ visual diffusion models to fuse social-account aliases and diversion numbers into the texture of the image itself. Because the resulting text no longer has the sharp boundaries of native fonts, it is highly camouflaged and difficult for conventional OCR systems to detect.\\
\textbf{[Conclusion]} Determination: \textbf{Violation (off-platform diversion)}.
\end{quote}

\subsubsection{Combination Camouflage Attack}
\begin{figure}[htbp]
    \centering
    \includegraphics[width=0.8\textwidth, height=0.35\textheight, keepaspectratio]{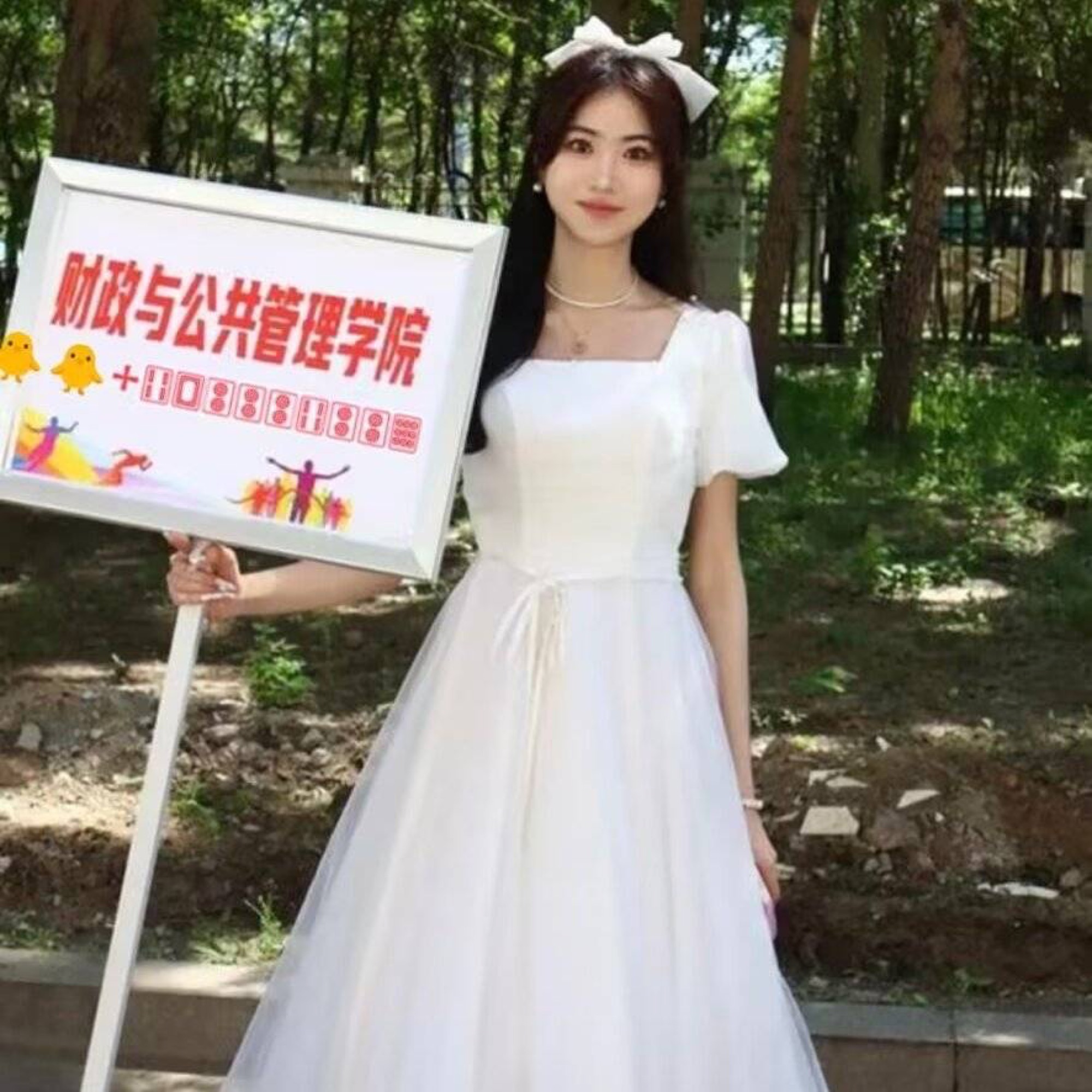}
    \caption{Combination-camouflage case: a normal academic background used to hide group numbers}
    \label{fig:case_combination}
\end{figure}
\begin{quote}
\textbf{[Observation]} The image uses a formal institutional or academic background, such as "School of Finance and Public Administration," as its overall visual framing. Beneath the text, a bird icon and a sequence of Mahjong tiles are carefully arranged.\\
\textbf{[Extraction]} Ignoring the misleading background, the model identifies the hidden contact information encoded with Mahjong-tile patterns below the school name and reconstructs the string "QQ+412224229".\\
\textbf{[Reasoning]} This is a typical multimodal combination-camouflage attack. The violator relies on authoritative scene semantics to create a seemingly safe context, while using image symbols for semantic indirection: the bird icon hints at QQ's penguin logo, and the Mahjong tiles are mapped to digits according to tile pattern and suit. This cross-modal conversion from object symbols to diversion numbers allows the attack to evade direct text detection by standard OCR systems.\\
\textbf{[Conclusion]} Determination: \textbf{Violation (off-platform diversion)}.
\end{quote}

\subsubsection{Long-Image Concealed-Text Attack}
\begin{figure}[p]
    \centering
    \includegraphics[width=0.95\textwidth, height=0.85\textheight, keepaspectratio]{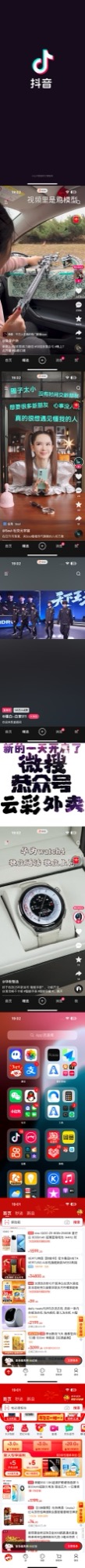}
    \caption{Long-image case: diversion text hidden beneath a routine greeting image}
    \label{fig:case_long}
\end{figure}
\begin{quote}
\textbf{[Observation]} The image is a daily "good morning" landscape poster with an extremely long vertical aspect ratio.\\
\textbf{[Extraction]} By tracing the long layout downward, the model recovers the sparsely distributed and structurally separated strings "微搜", "恭众号", and "云彩外卖"; the first two are homophonic variants commonly used to suggest "微信搜索公众号" ("search the public account on WeChat").\\
\textbf{[Reasoning]} This case exploits long-image concealed text. The attacker places normal, high-weight greeting text in the first visible screenful, then distributes homophonic variant characters related to diversion in the tail region that is often ignored after downsampling and cropping in industrial pipelines. Such fragmented, long-range cues are difficult to capture.\\
\textbf{[Conclusion]} Determination: \textbf{Violation (off-platform diversion)}.
\end{quote}
\end{CJK*}

\clearpage
\subsection{Evaluation Prompts}
\label{sec:appendix_prompts}

To ensure fair and objective evaluation, Xuanwu VL-2B uses standardized prompt configurations matched to each task type.

\subsubsection{Industrial Moderation Evaluation Templates}
During the evaluation of the industrial metrics in Tables~\ref{tab:business_moderation} and \ref{tab:business_spam_baseline}, we do not directly expose the original system prompts. Instead, we describe the evaluation tasks using ``category definition + structured template.'' The templates cover the following two settings:

\begin{quote}
\textbf{Content moderation (taking ad as an example)}: given an input sample $x$, the model outputs a binary label $y \in \{\text{Yes}, \text{No}\}$ according to a predefined ad classification criterion $\rightarrow$ label the sample as ``Yes'' when its main content serves commercial promotion, recruitment, transaction posting, or product/service marketing $\rightarrow$ label it as ``No'' when such information appears only as background or as an auxiliary element rather than the main focus.
\end{quote}

\begin{quote}
\textbf{Adversarial OCR}: \textbf{[Observation]} describes the main subjects and background of the image $\rightarrow$ \textbf{[Extraction]} identifies all visible or obscure text and symbols in the image $\rightarrow$ \textbf{[Reasoning]} compares the extracted evidence against moderation standards $\rightarrow$ \textbf{[Conclusion]} outputs the final determination (Safe / Violating-Category).
\end{quote}

All evaluations use greedy decoding (temperature = 0) with a maximum output length of 8,192 tokens.

\subsubsection{General Multimodal Evaluation Prompt}
During the evaluation of general leaderboards, following standard practice, we used the prompt format below and deployed the model via lmdeploy with \texttt{do\_sample} set to \texttt{False}:

\begin{quote}
Please carefully read the image and the following multiple-choice question. Analyze the visual information and select the most accurate option from the provided choices. Enclose your final answer letter (A, B, C, or D) within [].
\end{quote}

\subsubsection{Text-Only Evaluation Prompts}
For text-only benchmarks, we follow the community-standard prompts and answer-extraction rules used by each benchmark. For multiple-choice tasks, we preserve explicit answer slots or fixed answer formats; for generative tasks, we keep the original benchmark instruction format and do not inject business-specific system prompts.

%% file: sections/07-appendix-pretrain-en.tex
\subsection{Pre-Training Data Composition}
\label{sec:appendix_pretrain_data}

To supplement the high-level description of the Pre-Training stage in Section 3, Table~\ref{tab:appendix_pretrain_category_summary} summarizes the category-level composition of the 20,078,399 raw-source samples before they enter the pre-training pipeline, and Table~\ref{tab:appendix_pretrain_category_references} further consolidates the total sample counts and dataset references under the nine top-level categories. Importantly, the appendix reports the raw source-data scale, whereas the 1.3M, 17.33M, and 18.63M figures in Section 3 refer to the effective samples that actually participate in training after filtering, deduplication, and quality control. The two totals therefore use different accounting scopes. Overall, the pretraining corpus spans Captioning \& Knowledge, Chart \& Table, General VQA, Grounding \& Counting, Mathematics, Naive OCR, OCR QA, Science, and Text-only, balancing general knowledge acquisition with chart/table understanding, document and OCR competence, grounding/counting, as well as mathematical and scientific reasoning. For consistency, purely textual corpora are grouped into Text-only, while all remaining datasets are categorized by their primary task. For a small number of ambiguous datasets, we assign them based on their data content and primary task. When no corresponding paper is available, a public dataset URL is used as the reference source.

\begin{table}[ht]
\centering
\scriptsize
\caption{Category-level statistics of the raw source data for the Pretrain stage}
\label{tab:appendix_pretrain_category_summary}
\renewcommand{\arraystretch}{1.12}
\begin{tabular}{p{4.2cm}p{2.0cm}p{2.6cm}}
\hline
\textbf{Category} & \textbf{\#Datasets} & \textbf{\#Samples} \\ \hline
Captioning \& Knowledge & 13 & 3,956,963 \\
Chart \& Table & 16 & 1,968,589 \\
General VQA & 34 & 3,266,107 \\
Grounding \& Counting & 7 & 1,464,624 \\
Mathematics & 7 & 567,529 \\
Naive OCR & 26 & 2,232,976 \\
OCR QA & 16 & 785,379 \\
Science & 8 & 510,128 \\
Text-only & 16 & 5,326,104 \\
\hline
\end{tabular}
\end{table}

\begingroup
\scriptsize
\setlength{\tabcolsep}{3pt}
\renewcommand{\arraystretch}{1.12}
\setlength{\LTleft}{0pt}
\setlength{\LTright}{0pt}
\begin{longtable}{@{}>{\raggedright\arraybackslash}p{2.4cm}>{\raggedleft\arraybackslash}p{1.5cm}>{\raggedright\arraybackslash}p{10.2cm}@{}}
\caption{Top-level summary of the raw source data for the Pretrain stage and dataset references}
\label{tab:appendix_pretrain_category_references}\\
\hline
\textbf{Top-level Category} & \textbf{\#Samples} & \textbf{Datasets and References} \\ \hline
\endfirsthead
\caption[]{Top-level summary of the raw source data for the Pretrain stage and dataset references (continued)}\\
\hline
\textbf{Top-level Category} & \textbf{\#Samples} & \textbf{Datasets and References} \\ \hline
\endhead
\hline
\endfoot
\hline
\endlastfoot
Captioning \& Knowledge & 3,956,963 & DenseFusion-1M \cite{pretrain_densefusion_1m}; ShareGPT4V \cite{pretrain_sharegpt4v}; CC3M \cite{pretrain_cc3m}; ALLaVA (LAION-Caption) \cite{pretrain_allava}; Image-Textualization \cite{pretrain_image_textualization}; ALLaVA (VFLAN-Caption) \cite{pretrain_allava}; ShareGPT-4o \cite{pretrain_sharegpt_4o}; WikiArt \cite{pretrain_wikiart}; Movie-Posters \cite{pretrain_movie_posters_100k}; KVQA \cite{pretrain_kvqa}; TextCaps \cite{pretrain_textcaps}; TMDB-Celeb-10K \cite{pretrain_tmdb_celeb_10k}; Emo-Visual-Data \cite{pretrain_emo_visual_data} \\
Chart \& Table & 1,968,589 & UniChart \cite{pretrain_unichart_pretrain_data}; PlotQA \cite{pretrain_plotqa}; MMC-Inst \cite{pretrain_mmc_inst}; DVQA \cite{pretrain_dvqa}; FigureQA \cite{pretrain_figureqa}; Block-Diagram \cite{pretrain_block_diagram}; VQAonBD \cite{pretrain_vqaonbd}; MapQA \cite{pretrain_mapqa}; ChartQA \cite{pretrain_chartqa}; Chart2Text \cite{pretrain_chart2text}; InfoVQA \cite{pretrain_infovqa}; VisText \cite{pretrain_vistext}; MultiHiertt \cite{pretrain_multihiertt}; LRV-Instruction \cite{pretrain_lrv_instruction}; TAT-DQA \cite{pretrain_tat_dqa}; Diagram-Image-To-Text \cite{pretrain_diagram_image_to_text} \\
General VQA & 3,266,107 & ALLaVA (LAION-Instruct, Stage 1.5) \cite{pretrain_allava}; ALLaVA (LAION-Instruct, Stage 0.5) \cite{pretrain_allava}; MMInstruct-GPT4V \cite{pretrain_mminstruct_gpt4v}; LNQA \cite{pretrain_lnqa}; LVIS-Instruct4V \cite{pretrain_lvis_instruct4v}; ALLaVA (VFLAN-Instruct, Stage 1.5) \cite{pretrain_allava}; ALLaVA (VFLAN-Instruct, Stage 0.5) \cite{pretrain_allava}; LLaVA-Instruct-150K (ZH) \cite{pretrain_llava_en_zh_300k}; LLaVA-Instruct-150K (EN) \cite{pretrain_llava_en_zh_300k}; NLVR2 \cite{pretrain_nlvr2}; RLAIF-V \cite{pretrain_rlaif_v}; VQAv2 \cite{pretrain_vqav2}; COCO-QA \cite{pretrain_coco_qa}; LLaVA-Critic-113K (Pointwise) \cite{pretrain_llava_critic_113k}; GQA \cite{pretrain_gqa}; MIMIC-CGD \cite{pretrain_mimic_cgd}; DaTikZ \cite{pretrain_datikz}; LLaVA-Critic-113K (Pairwise) \cite{pretrain_llava_critic_113k}; KonIQ-10k \cite{pretrain_koniq_10k}; ICON-QA \cite{pretrain_icon_qa}; LLaVAR \cite{pretrain_llavar}; Places365 \cite{pretrain_places365_custom}; A-OKVQA \cite{pretrain_aokvqa}; IDK \cite{pretrain_idk}; LAION-GPT4V \cite{pretrain_laion_gpt4v}; WebSight \cite{pretrain_websight}; Hateful Memes \cite{kiela2020hatefulmemes}; Spot-the-Diff \cite{pretrain_spot_the_diff}; Memotion \cite{pretrain_memotion}; WildVision \cite{pretrain_wildvision}; SketchyVQA \cite{pretrain_sketchyvqa}; VizWiz \cite{pretrain_vizwiz_train}; Indoor Scene Classification \cite{pretrain_indoor_scene_classification}; DriveLM \cite{pretrain_drivelm} \\
Grounding \& Counting & 1,464,624 & Objects365 \cite{pretrain_objects365}; TallyQA \cite{pretrain_tallyqa}; SA-1B \cite{pretrain_sa_1b}; RefCOCO \cite{pretrain_refcoco}; RefCOCO+ \cite{pretrain_refcoco}; SpatialSense \cite{pretrain_spatialsense}; GroundUI-18K \cite{pretrain_groundui_18k} \\
Mathematics & 567,529 & MAVIS (Function) \cite{pretrain_mavis_function}; MAVIS (Geometry) \cite{pretrain_mavis_function}; TabMWP \cite{pretrain_tabmwp}; CLEVR-Math \cite{pretrain_clevr_math}; UniGeo \cite{pretrain_unigeo}; Geometry3K \cite{pretrain_geometry3k}; GeoS \cite{pretrain_geos} \\
Naive OCR & 2,232,976 & Latex-Formula \cite{pretrain_latex_formulas}; SynthDoG-EN \cite{pretrain_synthdog_en}; SynthDoG-ZH \cite{pretrain_synthdog_en}; K12-Printing \cite{pretrain_k12_printing}; Handwriting-Latex \cite{pretrain_handwriting_latex}; HME-100K \cite{pretrain_hme100k}; TAL-OCR-Composed-37K \cite{pretrain_tal_ocr_composed_37k}; SROIE \cite{pretrain_sroie}; LSVT \cite{pretrain_icdar2019_lsvt}; CTW1500 \cite{pretrain_ctw}; ReCTS \cite{pretrain_icdar2019_rects}; COCO-Text \cite{pretrain_coco_text}; Handwritten-Mathematical-Expression \cite{pretrain_handwritten_mathematical_expression_convert_latex}; CAPTCHA \cite{pretrain_captcha_dataset}; IAM \cite{pretrain_iam}; MTWI \cite{pretrain_mtwi}; Chrome-Writting \cite{pretrain_chrome_writting}; HierText OCR \cite{pretrain_ocr_hiertext}; RenderedText \cite{pretrain_renderedtext}; IMGUR5K \cite{pretrain_imgur5k}; ICDAR 2019 ArT \cite{pretrain_icdar2019_art}; WordArt \cite{pretrain_wordart}; Invoices and Receipts OCR v1 \cite{pretrain_invoices_and_receipts_ocr_v1}; ORAND-CAR \cite{pretrain_orand_car_a}; IIIT-5K \cite{pretrain_iiit5k}; SVRD \cite{pretrain_svrd} \\
OCR QA & 785,379 & Docmatix \cite{pretrain_docmatix}; OCR-VQA \cite{pretrain_ocr_vqa}; UReader-Instruction-1.0 \cite{pretrain_ureader_instruction_1_0}; ColPali \cite{pretrain_colpali_train_set}; ScreenQA \cite{pretrain_screenqa}; DocReason25K \cite{pretrain_docreason25k}; TextVQA \cite{pretrain_textvqa}; ST-VQA \cite{pretrain_st_vqa}; SQuAD-VQA \cite{pretrain_squad_vqa}; EST-VQA \cite{pretrain_est_vqa}; DocVQA \cite{pretrain_docvqa}; Sujet-Finance-QA-Vision-100K \cite{pretrain_sujet_finance_qa_vision_100k}; PDF-VQA \cite{pretrain_pdfvqa}; MTVQA \cite{pretrain_mtvqa_train}; SlideVQA \cite{pretrain_slidevqa}; SROIE \cite{pretrain_sroie} \\
Science & 510,128 & VisualWebInstruct \cite{pretrain_visualwebinstruct}; ArxivQA \cite{pretrain_arxivqa}; PathVQA \cite{pretrain_path_vqa}; ScienceQA \cite{pretrain_scienceqa}; TQA \cite{pretrain_tqa}; WeatherQA-SFT \cite{pretrain_weatherqa_sft}; SPARK \cite{pretrain_spark}; VQA-RAD \cite{pretrain_vqa_rad} \\
Text-only & 5,326,104 & OpenMathInstruct-1 \cite{pretrain_openmathinstruct_1}; Infinity-Instruct \cite{pretrain_infinity_instruct}; OpenOrca \cite{pretrain_orca}; NuminaMath-CoT \cite{pretrain_numinamath_cot}; UltraInteract-SFT \cite{pretrain_ultrainteract_sft}; MathInstruct \cite{pretrain_mathinstruct}; Orca-Math \cite{pretrain_orca_math}; InfinityMATH \cite{pretrain_infinitymath}; OpenHermes-2.5 \cite{pretrain_openhermes_2_5}; TableLLM \cite{pretrain_tablellm}; WizardLM Evol-Instruct 70K \cite{pretrain_wizardlm}; Code-Feedback \cite{pretrain_code_feedback}; MetaMathQA-40K \cite{pretrain_metamathqa_40k}; Python-Codes-25K \cite{pretrain_python_codes_25k}; Python-Code-Instructions-18K-Alpaca \cite{pretrain_python_code_instructions_18k_alpaca}; Math-Step-DPO-10K \cite{pretrain_math_step_dpo_10k} \\
\end{longtable}
\endgroup

\noindent\textit{Note:} Purely textual corpora are grouped into Text-only, while the remaining datasets are categorized by their primary task. When a dataset does not have a corresponding paper, the citation uses a public dataset URL instead. A small number of ambiguous datasets are grouped based on their data content and primary task.

\clearpage